\pdfoutput=1

\documentclass[11pt]{article}
\usepackage[]{acl}
\usepackage{times}
\usepackage{latexsym}
\usepackage[T1]{fontenc}
\usepackage[utf8]{inputenc}

\usepackage{microtype}
\usepackage{graphicx,amssymb}
\usepackage{multirow}
\usepackage{adjustbox}
\usepackage{booktabs}

\usepackage{bbm}

\usepackage{array}
\usepackage{bm}

\usepackage{soul}
\usepackage{color, xcolor}
\usepackage{pifont}
\definecolor{airforceblue}{rgb}{0.36, 0.54, 0.66}
\definecolor{blue(ncs)}{rgb}{0.0, 0.53, 0.74}
\definecolor{blush}{rgb}{0.87, 0.36, 0.51}
\definecolor{alizarin}{rgb}{0.82, 0.1, 0.26}
\definecolor{ao(english)}{rgb}{0.0, 0.5, 0.0}
\definecolor{hlgreen}{HTML}{B2D5CB}
\definecolor{hlblue}{HTML}{ADD8E6}
\definecolor{hlyellow}{HTML}{EADDCA}

\newcommand{\cmark}{\ding{51}}%
\newcommand{\xmark}{\ding{55}}%

\usepackage{CJKutf8}

\newcommand\extrafootertext[1]{%
    \bgroup
    \renewcommand\thefootnote{\fnsymbol{footnote}}%
    \renewcommand\thempfootnote{\fnsymbol{mpfootnote}}%
    \footnotetext[0]{#1}%
    \egroup
}

\title{
\emph{SeaEval} for Multilingual Foundation Models: \\ 
From Cross-Lingual Alignment to Cultural Reasoning
}

\author{
Bin Wang\textsuperscript{$1$*}, Zhengyuan Liu\textsuperscript{$1$*}, Xin Huang\textsuperscript{$1$}, Fangkai Jiao\textsuperscript{$2$}, Yang Ding\textsuperscript{$1$},
\\
\bf AiTi Aw\textsuperscript{$1$}, Nancy F. Chen\textsuperscript{$1,3$}\\
\textsuperscript{$1$}Institute for Infocomm Research (I$^2$R), A*STAR, Singapore\\
\textsuperscript{$2$}Nanyang Technological University, Singapore
\\
\textsuperscript{$3$}Centre for Frontier AI Research (CFAR), A*STAR, Singapore\\
\texttt{\{wang\_bin,liu\_zhengyuan,nfychen\}@i2r.a-star.edu.sg}
}

\begin{document}
\maketitle

\begin{abstract}
We present \textit{SeaEval}, a benchmark for multilingual foundation models. In addition to characterizing how these models understand and reason with natural language, we also investigate how well they comprehend cultural practices, nuances, and values.
Alongside standard accuracy metrics, we investigate the brittleness of foundation models in the dimensions of semantics and multilinguality. 
Our analyses span both open-sourced and closed models, leading to empirical results across classic NLP tasks, reasoning, and cultural comprehension. Key findings indicate 
(1) Many models exhibit varied behavior when given paraphrased instructions.
(2) Many models still suffer from exposure bias (e.g., positional bias, majority label bias).
(3) For questions rooted in factual, scientific, and commonsense knowledge, consistent responses are expected across multilingual queries that are semantically equivalent. Yet, most models surprisingly demonstrate inconsistent performance on these queries.
(4) Multilingually-trained models have not attained ``balanced multilingual'' capabilities. Our endeavors underscore the need for more generalizable semantic representations and enhanced multilingual contextualization.
\textit{SeaEval} can serve as a launchpad for more thorough investigations and evaluations for multilingual and multicultural scenarios.\footnote{ Datasets, evaluation toolkit, and leaderboard are available at \url{https://github.com/SeaEval/SeaEval}.}\extrafootertext{{*}: Equal contribution.}
    
\end{abstract}

\section{Introduction}

    Over the past years, there has been rapid development of large language models (LLMs), also known as a type of foundation models (FMs)~\cite{Bommasani2021FoundationModels}, demonstrating their generalizability and adaptability across various downstream tasks~\cite{scao2022bloom,chowdhery2022palm,openai2023gpt,touvron2023llama,wang-etal-2023-instructive}. The proliferation of LLMs has raised urgent requirements for extensively evaluating their performance in various contexts, and understanding their limitations \cite{wei2023overview}. Therefore, recent efforts on LLM evaluation are focusing on more challenging and human-centric tasks including complex reasoning ~\cite{clark2018think,zellers2019hellaswag,hendrycks2021measuring} and domain-knowledge-intensive problems~\cite{hendrycks2021measuring,lin-etal-2022-truthfulqa,zhong2023agieval,srivastava2023beyond}, math problems~\cite{zhong2023agieval}, human exams~\cite{clark2018think,zhong2023agieval,openai2023gpt}, and using LLMs as judges for open question answering~\cite{zheng2023judging}. 

    \begin{figure}[t]
        \centering
         \includegraphics[width=0.4\textwidth]{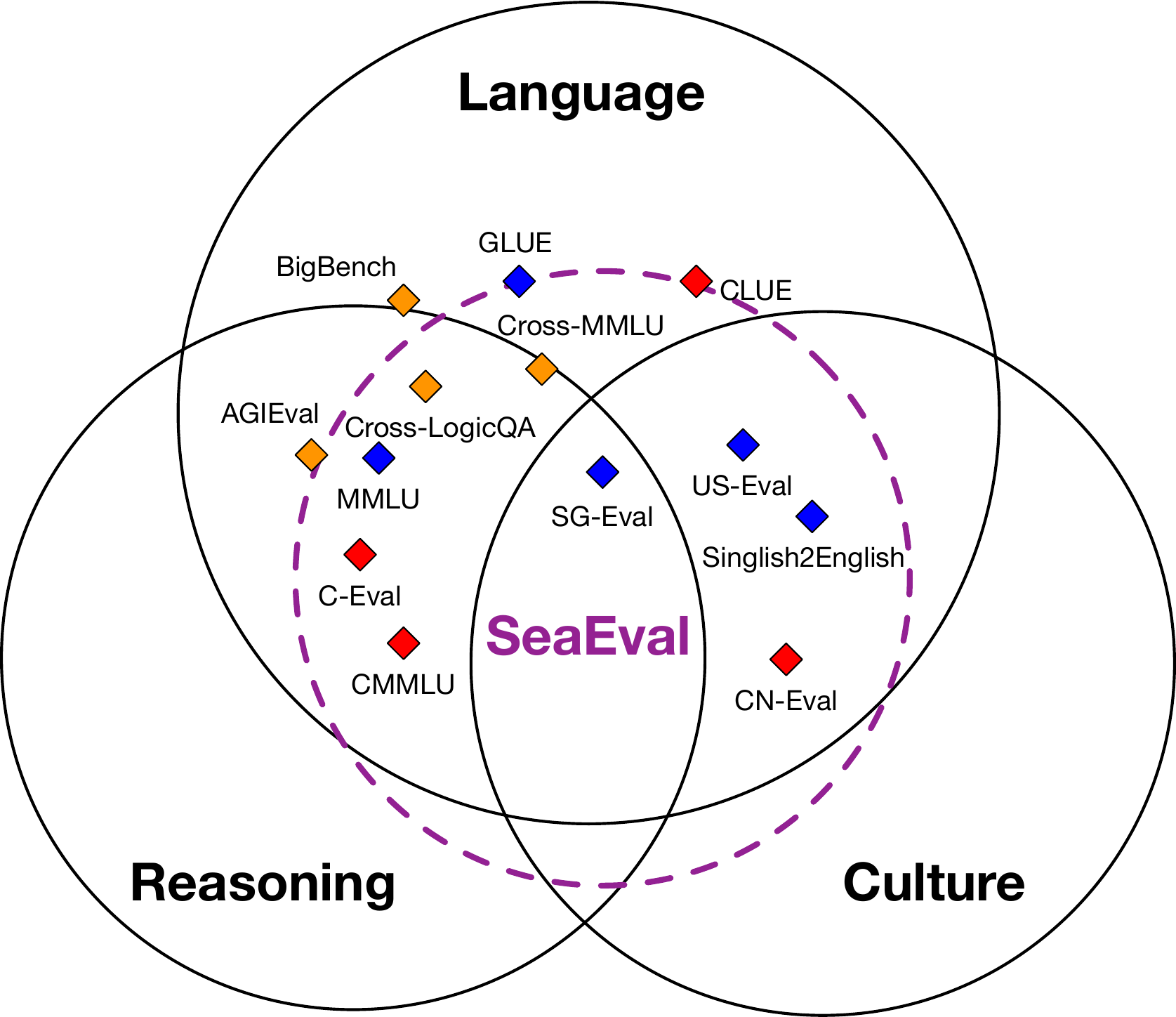}
        \caption{\textit{SeaEval} for multilingual foundation models. English is represented by the color blue, Chinese by red, and a mix of multiple languages by yellow. \textit{SeaEval} includes the datasets within the dotted-line circle.}
        \label{fig:capability_chart}
    \vspace{-0.2cm}
    \end{figure}

    Compared to other species on earth, humans are adept in using language to symbolize and encode thoughts, represent and document knowledge, and express emotions. Whether it is spoken or written form, language has become a default means for humans to conduct and communicate our reasoning process and logic \cite{logan1986alphabet,li2002turning}. Three variables are language, region, and culture. Our cultural behavior, rituals, and values are often embodied and represented through language \cite{kramsch1991culture,ji2004culture}. Therefore, the capabilities of language models go beyond language per se. Evaluating multilingual language model should incorporate deep cultural understanding and reasoning. To this end, we expand the current evaluation criteria to cover more linguistic and cultural contexts \cite{ahuja2023mega,lai2023chatgpt}, which are under-explored in prior research.

    Another important yet under-explored aspect of evaluating multilingual foundation models is their knowledge transferability in language dimensions. Multilingual foundation models are expected to demonstrate consistent performance across languages in the context of region-invariant common knowledge~\cite{zhu2023extrapolating}. Monolingual or bilingual benchmarks cannot adequately capture this aspect. Therefore, we introduce cross-lingual consistency evaluation with tailored datasets and specialized metrics.
    
    \textit{SeaEval} aims to assess the capabilities of multilingual foundation models in four dimensions: (a) Classic NLP tasks that are centered around language understanding and generation; (b) Complex reasoning; (c) Cultural understanding and reasoning; and (d) Cross-lingual knowledge transfer and contextualization. \textit{SeaEval} encompasses a total of 28 datasets, including 6 new datasets constructed for cultural reasoning and cross-lingual consistency assessments.

    Key findings from our investigations and experimental results indicate: (1) Most models show varying responses to rephrased instructions. (2) Exposure bias (e.g., positional bias, majority label bias) of label arrangements still prevails. (3) Most models give inconsistent answers when the same fact-based questions are asked in different languages. This counter-intuitive observation suggests semantic representations are not generalized in the multilingual context. (4) Multilingually-trained models fall short of achieving ``balanced multilingual'' proficiency.
    To sum up, our contributions are multifold: 
    \begin{itemize}
        \item We offer fresh insights into multilingual foundation models and their evaluations.
    
        \item We introduce 7 new datasets for assessing cultural understanding and cross-lingual consistency, along with tailored metrics to fill existing gaps in model evaluation.
        
        \item We present a comprehensive evaluation benchmark derived from extensive experiments across models, tasks, datasets and metrics. This framework facilitates in-depth exploration of multilingual and multicultural tasks using foundation models.

\end{itemize}

    \begin{table*}[t]
        \centering
        \begin{adjustbox}{width=1.0\textwidth,center}
        \begin{tabular}{ l | p{5.5cm} | p{5.5cm} | p{5.5cm} | p{5.5cm} }
        \toprule
         \textbf{Language} & \multicolumn{1}{c|}{\textbf{English}} & \multicolumn{1}{c|}{\textbf{Chinese}} & \multicolumn{1}{c|}{\textbf{Indonesian}} & \multicolumn{1}{c}{\textbf{Spanish}}
         \\ \midrule
         
         \textbf{Question}
         & \emph{Please choose the correct answer for} 
         & \begin{CJK*}{UTF8}{gbsn} 回答下面问题，选择正确答案。\end{CJK*}
         & \emph{Silakan pilih jawaban yang benar}
         & \emph{Por favor elija la respuesta correcta}
         \\ 
         
         & \emph{the following question.} 
         & \begin{CJK*}{UTF8}{gbsn} 当白光通过棱镜时，比绿光弯曲 \end{CJK*} 
         & \emph{untuk pertanyaan berikut.} 
         & \emph{para la siguiente pregunta.}
         \\
         
         & \emph{When white light passes through a } 
         & \begin{CJK*}{UTF8}{gbsn} 更多的光是？ \end{CJK*}
         & \emph{Ketika cahaya putih melewati sebuat}
         & \emph{Cuando la luz blanca pasa a través}
         \\
         
         & \emph{prism, the light that bends more than} 
         & \begin{CJK*}{UTF8}{gbsn} (A) 红色的 \end{CJK*}
         & \emph{prisma, cahaya manakah yang}
         & \emph{de un prisma, la luz que se desvía}
         \\
         
         & \emph{green is?} 
         & \begin{CJK*}{UTF8}{gbsn} (B) 黄色的 \end{CJK*}
         & \emph{memiliki sudut deviasi lebih besar}
         & \emph{más que la verde es}
         \\
         
         & \emph{(A) Red} 
         & \begin{CJK*}{UTF8}{gbsn} (C) 蓝色的 \end{CJK*}
         & \emph{daripada cahaya hijau?}
         & \emph{(A) Rojo}
         \\
         
         & \emph{(B) Yellow} 
         & \begin{CJK*}{UTF8}{gbsn} (D) 都不是 \end{CJK*}
         & \emph{(A) Merah}
         & \emph{(B) Amarillo}
         \\
         
         & \emph{(C) Blue} 
         & 
         & \emph{(B) Kuning}
         & \emph{(C) Azul}
         \\

         & \emph{(D) None of these}
         &
         & \emph{(C) Biru}
         & \emph{(D) Ninguna de las anteriores}
         \\ 

         &
         &
         & \emph{(D) Tak ada satupun}
         &
         \\
         \midrule
         
         \textbf{Answer} 
         & \emph{The correct answer is:}
         & \begin{CJK*}{UTF8}{gbsn} (A) 红色的 \end{CJK*}
         & \emph{(D) Tak ada satupun}
         & \emph{La respuesta correcta es:}
         \\

         & \emph{(C) Blue }
         & \emph{\textbf{In English}: (A) Red }
         & \emph{\textbf{In English}: (D) None of them }
         & \emph{(A) Rojo } \emph{\textbf{In English}: (A) Red}
         \\

         \midrule

         \textbf{Correctness}
         & \multicolumn{1}{c|}{\textcolor{ao(english)}{\cmark}} 
         & \multicolumn{1}{c|}{\textcolor{red}{\xmark}} 
         & \multicolumn{1}{c|}{\textcolor{red}{\xmark}}
         & \multicolumn{1}{c}{\textcolor{red}{\xmark}}
         \\
         \bottomrule
        \end{tabular}
        \end{adjustbox}
        \caption{
            An example from Cross-MMLU dataset for evaluating cross-lingual consistency. Outputs are from ChatGPT. The answers are inconsistent for the same question posed in different languages. This inconsistency highlights insufficient alignment across languages, leading to suboptimal multilingual contextualization and representations.
        }

        \label{tab:cross-lingual-alignment-example}
    \end{table*}

\section{Essential Properties of Multilingual Foundation Models and Benchmarks}

    In this section, we delve into the desired properties of multilingual foundation models and explore the ideology for crafting a comprehensive benchmark.

    \subsection{Multilingual Foundation Model}

        Multilingual foundation models should possess additional properties beyond monolingual models to effectively handle diverse languages and cultural contexts. Here are the key properties:

        \textbf{Multilinguality} The applicability of multilingual LLMs can be elevated when they demonstrate proficiency across diverse languages, including low-resource languages and their dialects \cite{wu2016google,kulshreshtha-etal-2020-cross}. The advantage of multilingualism is that it allows these models to bridge the linguistic gaps that exist between different cultures and communities. Multilingual capability is not a combination of various monolingual capabilities \cite{ye2023language}; instead, it represents a holistic approach to understanding and processing languages. For example, the model should be adept at bilingual tasks such as machine translation and code-switching scenarios, which offers the advantage of preserving linguistic and cultural information.

        \textbf{Reasoning Capability}
        Reasoning has long been treated as a complex yet essential capability in cognition that goes beyond fundamental language understanding \cite{McCarthy1989AI}, which can be demonstrated via extracting critical information from the text environment and drawing correct conclusions \cite{nilsson1991logic}. The advancement of natural language reasoning has evolved from explicit and superficial reading comprehension and natural language inference \cite{rajpurkar-etal-2016-squad,wang2018glue,wang-li-2023-relational} to encompass implicit, complex, and specific reasoning capabilities such as multi-hop reasoning, numerical reasoning, and logical reasoning \cite{race,cosmosqa,jiao2022merit}. While reasoning can be challenging due to the various relations and expressions involved, which are difficult to transcribe into symbolic or formal languages, FMs are shown to serve as a proxy to compress abundant knowledge, and solve various tasks following human instructions with less specialization \cite{cot,voyager,jiao2023logicllm}.

        \textbf{Cultural Understanding} Language is deeply tied to culture and local norms~\cite{pennycook2006global}. The meaning of linguistic elements can differ considerably across cultures. Cultural understanding capability can help large language models better interpret content with local communication conventions \cite{zampieri2020natural} and avoid stereotypes and biases. 
        In the context of philosophy, language, reasoning, and culture are considered three important pillars that play a significant role in shaping people's understanding of the world~\cite{ji2004culture,kramsch2014language} as depicted in Figure~\ref{fig:capability_chart}. They intersect and influence one another in various ways across studies in linguistics, philosophy, and psychology. \citet{logan1986alphabet} presents a provocative proposal that language can be used to account for cultural differences in reasoning styles. 
        Therefore, in pursuit of advancing multilingual foundation models, it is desired that models not only acquire proficiency across languages but also gain a profound comprehension of cultural concepts influencing human behaviors.
        An illustrative example highlighting the impact of local cultural conventions is shown in Table~\ref{tab:sample_sg_eval}, where a particular model must draw insights from locally sourced content to address this problem properly.

        \begin{table}[t]
            \centering
            \begin{adjustbox}{width=0.5\textwidth,center}
            \begin{tabular}{ l | l }
            \toprule
             \textbf{Question} & \emph{Which drink in Singapore has the highest calories?} \\ 
             & \emph{(A) Teh O} \\
             & (B) \emph{Teh Siew Dai} \\
             & (C) \emph{Kopi} \\
             & (D) \emph{Kopi C} \\
             \midrule
             \textbf{Multicultural} & \textcolor{alizarin}{\textbf{Multilingual Understanding}} \\
             \textbf{Reasoning Steps} & (Hokkien) Teh = Tea \\
             & (Cantonese) Siew Dai = Less Sweet/Sugar \\
             & (Malay) Kopi = Coffee \\
             & \textcolor{alizarin}{\textbf{Cultural/Personal Preferences}} \\
             & Teh = Tea + Condensed Milk + Sugar \\
             & Teh O = Tea + Sugar \\
             & Kopi = Coffee + Condensed Milk \\
             & Kopi C = Coffee + Evaporated Milk + Sugar \\
             & \textcolor{alizarin}{\textbf{Reasoning with Dietary Knowledge}} \\
             & Condensed milk = Sweetened = Sugar was Added\\
             & Sugar = Calories \\
             & Pure Tea or Coffee = Almost No Calories \\ \midrule
             \textbf{Answer} & (C) Kopi \\
        
            \bottomrule
            \end{tabular}
            \end{adjustbox}
            \caption{
                An example from SG-Eval dataset. To accomplish the task, one needs to employ reasoning that incorporates multilingual and cultural knowledge.
            }
            \label{tab:sample_sg_eval}
        \end{table}
    
        \textbf{Cross-Lingual Knowledge Transfer}
        An important advantage of encompassing multiple languages is the ability to access information from various language resources simultaneously, a characteristic that is also desired in multilingual foundation models. An effective cross-lingual knowledge transfer method can significantly enhance model capabilities across all languages, as they mutually reinforce each other. Additionally, world knowledge is typically dispersed in various languages and regions and may not be easily accessible in a single source, which demonstrates the need for cross-lingual knowledge transfer. On the other hand, some world knowledge should be kept consistent across different languages, such as factual, scientific, and commonsense knowledge. In Table~\ref{tab:cross-lingual-alignment-example}, we see an illustration of the same question posed in 4 languages, revealing inconsistent answers attributed to inadequate cross-lingual alignments of multilingual foundation models. Up to 16 languages are tested to illustrate this phenomenon as depicted in Section~\ref{appendix:more_examples}.

    \subsection{Multilingual Benchmarks}

        Motivated by the preceding discussion regarding the desired model characteristics, we introduce the targeted aspects for benchmarks:
        
        \textbf{Monolingual and Cross-lingual Capabilities}
        The focus on monolingual tasks ensures the model's proficiency in comprehending and generating text within a single language. The cross-lingual tasks, such as machine translation and code-switch comprehension, can access the communication capabilities across different languages, reflecting a comprehensive understanding of multilingual contexts. In terms of evaluation aspects, both fundamental NLP capabilities and complex reasoning capabilities should be examined under monolingual and cross-lingual settings.
        
        \textbf{Knowledge Transfer Ability}
        Language-related knowledge can be categorized into: 1) cultural knowledge and local norms tied to language and 2) common (universal) knowledge. 
        Cultural knowledge refers to language-related information that is specific to a particular culture, community, or region. It includes the nuances, customs, and norms associated with language use within a specific cultural context. Common knowledge is widely applicable across languages and communities, encompassing factual, scientific, and real-world knowledge, etc \cite{hendrycks2021measuring}.
        When designing evaluation benchmarks, it is essential to include a diverse set of language-related cultural tasks while also evaluating how effectively the universal knowledge is shared across different languages. Since such datasets are not readily accessible, this evaluation aspect is severely constrained in existing benchmarks.

        \textbf{Robustness and Stability}
        The robust context modeling and stable output generation are important to ensure LLMs work as intended (functionality) when applied to real-world applications (reliability) \cite{haduong2023risks}.
        When built on the auto-regressive framework, language models are originally trained to predict the next token given a sequence of previous ones, and their in-context learning and zero-shot inference performance depends on the prompts they receive \cite{ouyang2022training}. Consequently, minor variations of the input can possibly lead to distinct outputs with unpredictable formats. In particular, since FMs do not attain ``balanced multilingual'' capabilities, they are more sensitive to input variations such as multilingual and code-switch under real-world scenarios. Therefore, recognizing the models' instruction sensitivity should be a crucial aspect of the evaluation framework.

    \begin{table*}[t]
        \centering
        \begin{adjustbox}{width=1.00\textwidth,center}
        \begin{tabular}{ p{6.5cm} | c | c | p{1.7cm}<{\centering} | c  }
        \toprule
        
        \textbf{Dataset} & \textbf{Task Description} & \textbf{Languages} & \textbf{Metrics} & \textbf{\# of Samples} \\ \hline \hline
        
        \multicolumn{5}{l}
        {\sethlcolor{hlgreen}\hl{\textbf{Multicultural and Multilingual Understanding}}}
        \\ \hline \hline
        SG-Eval\textsuperscript{$\blacktriangle$} & Cultural Understanding & Eng & Accuracy & 102 \\
        US-Eval\textsuperscript{$\blacktriangle$} & Cultural Understanding & Eng & Accuracy & 102 \\
        CN-Eval\textsuperscript{$\blacktriangle$} & Cultural Understanding & Zho & Accuracy & 105 \\
        PH-Eval\textsuperscript{$\blacktriangle$} & Cultural Understanding & Eng & Accuracy & 100 \\
        Singlish2English\textsuperscript{$\blacktriangle$} & Multilingual Translation & Eng, Singlish & BLEU & 546 \\ \hline\hline
        \multicolumn{5}{l}
        {\sethlcolor{pink}\hl{\textbf{Cross-Lingual Consistency}}}
        \\ \hline\hline
        \multirow{1}{*}{Cross-MMLU\textsuperscript{$\blacktriangle$}} & \multirow{1}{*}{Reasoning} & Eng, Zho, Ind, Spa, Vie, Zsm, Pil & AC3 & \multirow{1}{*}{900} \\

        \multirow{1}{*}{Cross-LogiQA\textsuperscript{$\blacktriangle$}} & \multirow{1}{*}{Logic Reasoning} & Eng, Zho, Ind, Spa, Vie, Zsm, Pil & AC3 & \multirow{1}{*}{1,056}\\

        \hline\hline
        
        \multicolumn{5}{l}
        {\sethlcolor{hlblue}\hl{\textbf{Complex Reasoning}}}
        \\ \hline\hline
        MMLU~\cite{hendryckstest2021} & Mixed Knowledge & Eng & Accuracy & 857 \\
        C-Eval~\cite{sun-etal-2019-dream} & Subject Knowledge & Zho & Accuracy & 1,346 \\
        CMMLU~\cite{li2023cmmlu} & Subject Knowledge & Zho & Accuracy & 280 \\
        ZBench~\cite{zbench} & Subject Knowledge & Zho & Accuracy & 33 \\
        \hline\hline

        \multicolumn{5}{l}
        {\sethlcolor{hlyellow}\hl{\textbf{Classic NLP Tasks}}}
        \\ \hline\hline
        FLoRes-Lang2eng~\cite{guzman-etal-2019-flores} & Translation, Bilingual & Ind, Vie, Zho, Zsm, Eng & BLEU & 3,988 \\

        Ind-Emotion~\cite{saputri2018emotion} & Sentiment Analysis & Ind & Accuracy & 300 \\
        OCNLI~\cite{hu-etal-2020-ocnli} & Textual Entailment & Zho & Accuracy & 300 \\
        C3~\cite{sun-etal-2020-investigating} & Reading Comprehension & Zho & Accuracy & 300 \\
        SAMSum~\cite{gliwa-etal-2019-samsum} & Summarization & Eng & ROUGE & 300 \\
        DialogSum~\cite{chen-etal-2021-dialogsum} & Summarization & Eng & ROUGE & 300 \\
        DREAM~\cite{sun-etal-2019-dream} & Dialogue Comprehension & Eng & Accuracy & 300 \\ 
        8 GLUE Tasks~\cite{wang2018glue} & Fundamental NLP & Eng & Accuracy & 2,148\\
        \hline\hline
        \multicolumn{1}{c|}{29 Datasets} & Mixed & 8 & Mixed & 13,263 \\
        
        \bottomrule
        \end{tabular}
        \end{adjustbox}
        \caption{
            Datasets from \textbf{SeaEval}. 
            Language abbreviations are from ISO 639-3 standard, where Eng, Zho, Ind, Spa, Vie, Zsm, and Fil indicate English, Chinese (Mandarin), Indonesian, Spanish, Vietnamese, Malay (Malaysian), and Filipino, respectively. Examples from our newly collected datasets ($\blacktriangle$) are shown in Table~\ref{tab:seaeval_examples}.
        }   
        \label{tab:seaeval_datasets}
    \end{table*}

\section{SeaEval}
    In this section, we present our \textit{SeaEval} benchmark from task selection, data curation, to evaluation protocols.
    Besides the evaluation of fundamental capabilities and complex reasoning, we also include the evaluation tasks on cultural understanding and cross-lingual alignment. The datasets are summarized in Table~\ref{tab:seaeval_datasets}.

    \subsection{Task Selection}
    
        \textbf{Fundamental Language Capabilities}
            The fundamental capabilities can be evaluated by a combination of classic NLP tasks of language understanding and generation. To ensure the diversity regarding both task and language, we collected 18 representative datasets from 5 languages.
            Previous studies \cite{shi2022language,ye2023language} show that English-centric LLMs demonstrate certain multilingual transfer ability, where the skills learned from one source language can be readily transferred to other languages. Therefore, for discriminative tasks, we select 8 tasks from the GLUE benchmark \cite{wang2018glue},  
            including SST-2, COLA, QQP, QNLI, MNLI, WNLI, RTE, and MRPC. 
            Furthermore, we incorporate DREAM for English dialogue comprehension, OCNLI and C3 for Chinese comprehension, and Indo-Emotion dataset~\cite{saputri2018emotion,wilie-etal-2020-indonlu} to gauge emotion comprehension in Indonesian.
            To build a generative task basis, we include translation and summarization datasets from FLoRes, SAMSum, and DialogSum. 
    
        \textbf{Complex Reasoning}
            Classic NLP benchmarks (e.g., GLUE, SQuAD) primarily focus on text understanding rather than complex reasoning abilities aligned with intricate real-world scenarios. As language models continue to grow in size and complexity, it becomes increasingly important to assess their abilities in performing complex reasoning and problem-solving tasks that humans typically excel at~\citep{word-model}. Therefore, here we add evaluation datasets from recent representative human-centric benchmarks, which are derived from high-standard and professional exams. We include the MMLU dataset to assess knowledge comprehension in English. 
            To assess the reasoning capability in a multilingual setting, we include C-Eval, CMMLU, and ZBench, which are specifically tailored for evaluating intricate reasoning in Chinese.

        \textbf{Multilingual and Cultural Understanding}
            An effective multilingual language model is trained with text corpus from diverse sources. It enables the model to acquire cultural knowledge related to languages, which is important when serving users from different cultural backgrounds. In order to assess the model's cultural comprehension abilities, we manually construct 4 datasets containing multiple-choice questions that encompass 4 distinct regions: the United States (English), Singapore (English), China (Chinese), and the Philipines (English). The corresponding datasets are US-Eval, SG-Eval, CN-Eval, PH-Eval.
            Unlike monolingual models, multilingual models should demonstrate a strong capability for effectively transferring common knowledge. Therefore, we introduce two datasets, Cross-MMLU and Cross-LogiQA, to evaluate this feature across 7 diverse languages: English, Chinese, Indonesian, Spanish, Vietnamese, Malay, and Filipino.

    \subsection{Data Curation}

        Considering the size of LLMs, evaluation on the full test set can incur significant computational and economic expenses. Therefore, for existing datasets on evaluating model's fundamental capability and complex reasoning, we randomly sampled a subset. The numbers are listed in Table~\ref{tab:seaeval_datasets}, which results in over 13k samples in total.

        The output formats of autoregressive language models (e.g., GPT) cannot be easily controlled for open-ended tasks, making it difficult to assess the accuracy of their predictions. Consequently, in order to quantitatively evaluate their performance, we have transformed all the discriminative datasets (e.g., emotion classification, natural language inference, dialogue comprehension) into multiple-choice questions. While, for generative tasks such as summarization and translation, we have retained the original evaluation process, as it relies on word-matching metrics and human-annotated references are readily applicable.

        \textbf{Cultural Reasoning} There are no publicly available datasets for explicitly evaluating cultural knowledge in different regions. To effectively evaluate such knowledge, certain criteria should be met. First, the knowledge should originate directly from respective regions, distinct from widespread content. Second, it should encompass an understanding of the intricate norms of each culture under examination. Third, certain cultural expressions can be challenging to fully convey in another language, making it preferable to retain the knowledge in its original language.

        Therefore, we hired linguistic experts to construct datasets to evaluate the knowledge from three distinct regions, including the United States (US-Eval), Singapore (SG-Eval), China (CN-Eval) and the Philipines (PH-Eval). For each dataset, over 100 questions are sourced from a variety of channels, including local residencies' proposals, government websites, historical textbooks and exams, local cultural heritage materials, and pre-existing academic research datasets. CN-Eval and US-Eval also include questions carefully selected from MMLU, C-Eval and CMMLU datasets.
        Meanwhile, Singapore serves as an exceptional illustration, blending a harmonious fusion of diverse Southeast Asian cultures, enriched by a wealth of local content~\cite{deterding2007singapore,liu2022singlish,wang2024craft}. We also introduce a new dataset for Singlish to standard English translation with 546 sentences. Singlish incorporates elements of various languages, including Malay, Chinese dialects, and Tamil, and often includes unique vocabulary, grammar, and pronunciation. It has distinct local characteristics and requires a deep understanding of local practices. The samples from each dataset are illustrated in Table~\ref{tab:seaeval_examples}.

        \textbf{Cross-Lingual Consistency} As shown in Table~\ref{tab:cross-lingual-alignment-example}, for existing multilingual LLMs, the same question posed in different languages leads to inconsistent answers, which is undesired for multilingual foundation models. To qualitatively evaluate the model's capability in cross-lingual consistency, we present two datasets: Cross-MMLU and Cross-LogiQA with paralleled questions in 7 languages: English, Chinese, Indonesian, Spanish, Vietnamese, Malay, and Filipino. The selected questions are carefully curated to test common knowledge (e.g. commonsense, scientific), which is universally acceptable and transferrable between languages. Cross-MMLU and Cross-LogiQA are originated from MMLU dataset~\cite{hendrycks2021measuring} and LogiQA2.0 dataset~\cite{logic_qa}, respectively. To prepare questions that do not have equivalents in the target language, we utilize Google Translate first and enlist native speakers to perform proofreading and editing. This approach helps prevent translation errors and ensures accurate expressions, avoiding any potential misinterpretations.

        \begin{figure}[t]
            \centering
        \includegraphics[width=0.45\textwidth]{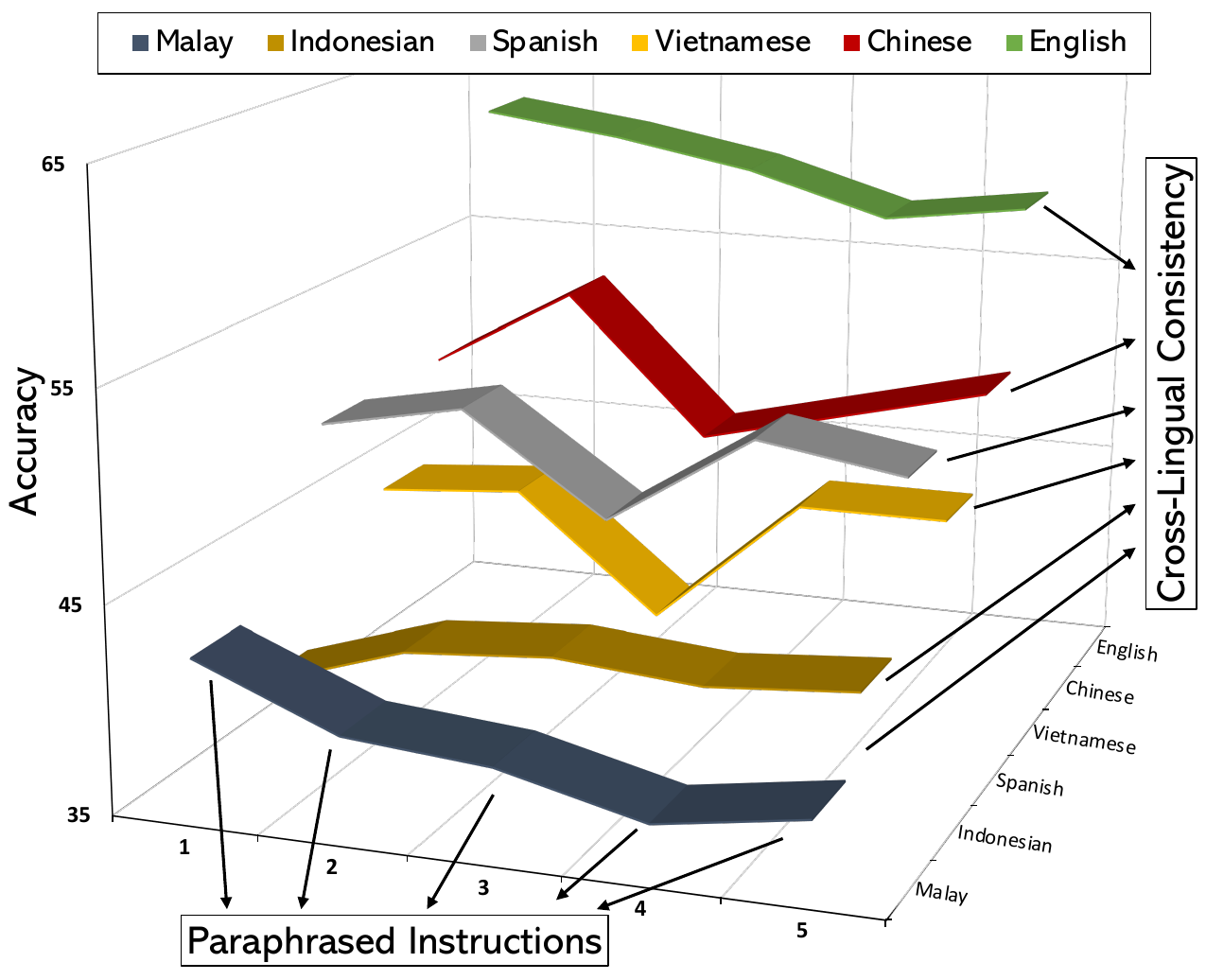}
            \caption{Two new evaluation protocols for multilingual foundation models in \emph{SeaEval}. The performance result is taken from ChatGPT on Cross-LogiQA dataset.}
        \label{fig:two_versa}
        \end{figure}

    \subsection{Evaluation Protocols}
    \label{sec:evaluation_protocols}
    
        Conventional benchmarks typically emphasize a single metric evaluation per dataset. As illustrated in Figure~\ref{fig:two_versa}, it becomes apparent that they do not provide enough coverage in multilingual FM evaluation. Therefore, in addition to standard metrics, we introduce two new evaluation dimensions called instruction sensitivity and cross-lingual consistency to measure a model's stability across instructions and languages.
        Regarding standard evaluation metrics, we use accuracy scores for multiple-choice questions. In the case of translation assessments, we report the BLEU-4 score~\cite{papineni-etal-2002-bleu}, while for summarization tasks, we deploy the average of ROUGE-1/2/L scores~\cite{lin-2004-rouge}.

        \textbf{Cross-Lingual Consistency} Besides the standard \emph{Accuracy} metric for evaluating multi-choice questions, we compute the cross-lingual \emph{Consistency} score as a measurement of whether the answers are consistent for the same question in 7 different languages without considering the answer's correctness. Specifically, for a question set $Q=\{q^1, q^2,...,q^N\}$, each question $q^i$ is represented in 7 languages $q^i=\{q^i_{eng},q^i_{zho},q^i_{ind},q^i_{spa},q^i_{vie},q^i_{msa},q^i_{fil}\}$, and $a^i_{lang}$ is model's answer to $q^i_{lang}$, the \emph{Consistency} score is computed as 
        $$M_{\{l_1,l_2,..,l_s\}}=\frac{\sum_{i=1}^{N}\mathbbm{1}_{\{a^i_{l_1}=a^i_{l_2}=..=a^i_{l_s}\}}}{N}$$ 
        $$Consistency_{s}=\frac{\sum_{\{l_1,l_2,..,l_s\} \in C(s,q_i)}M_{\{l_1,l_2,..,l_s\}}}{C_7^s}$$ 
        where $s\in[2,7]$. It measures the answer's consistency of any combination of $s$ languages. The model gets rewarded if it generates consistent answers across the sampled languages. The consistency requirement is enhanced to more languages with increased $s$. Given that both \textit{Accuracy} and \textit{Consistency} alone do not provide a comprehensive assessment of models' performance on cross-lingual datasets, we introduce the \emph{AC3} score as a holistic measure, which is calculated as the harmonic mean of both scores:
        $$AC3_s=2\cdot\frac{Accuracy\cdot Consistency_{s}}{Accuracy+Consistency_{s}}$$ where \emph{AC3} is within range $[0,1]$. We deploy \emph{AC3} with $s=3$ as the default value for Cross-MMLU and Cross-LogiQA datasets. Figure~\ref{fig:cross_consis_s_mmlu} illustrates the impact on variable $s$.
    
        \begin{figure}[t]
            \centering
             \includegraphics[width=0.49\textwidth]{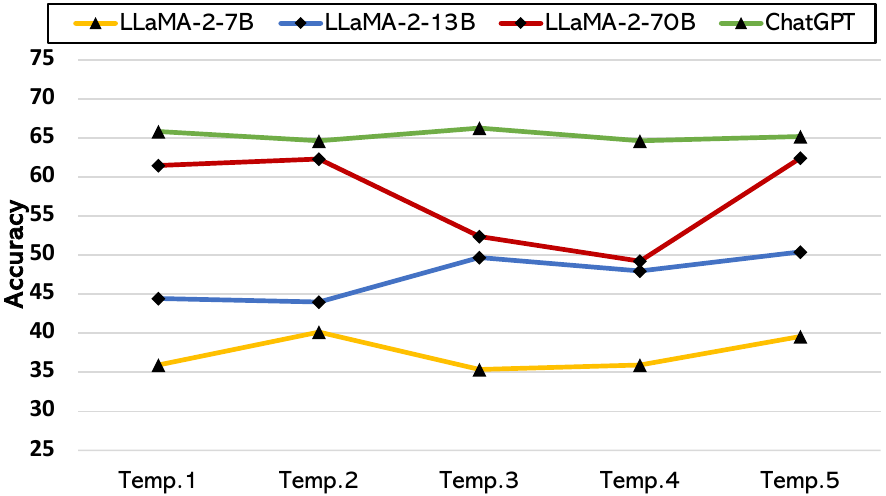}
            \caption{Performance on MMLU dataset with paraphrased instruction. Some models show large performance variances with paraphrased instruction templates.}
            \label{fig:instruction_templates}
        \end{figure}

        \begin{figure}[t]
            \centering
             \includegraphics[width=0.49\textwidth]{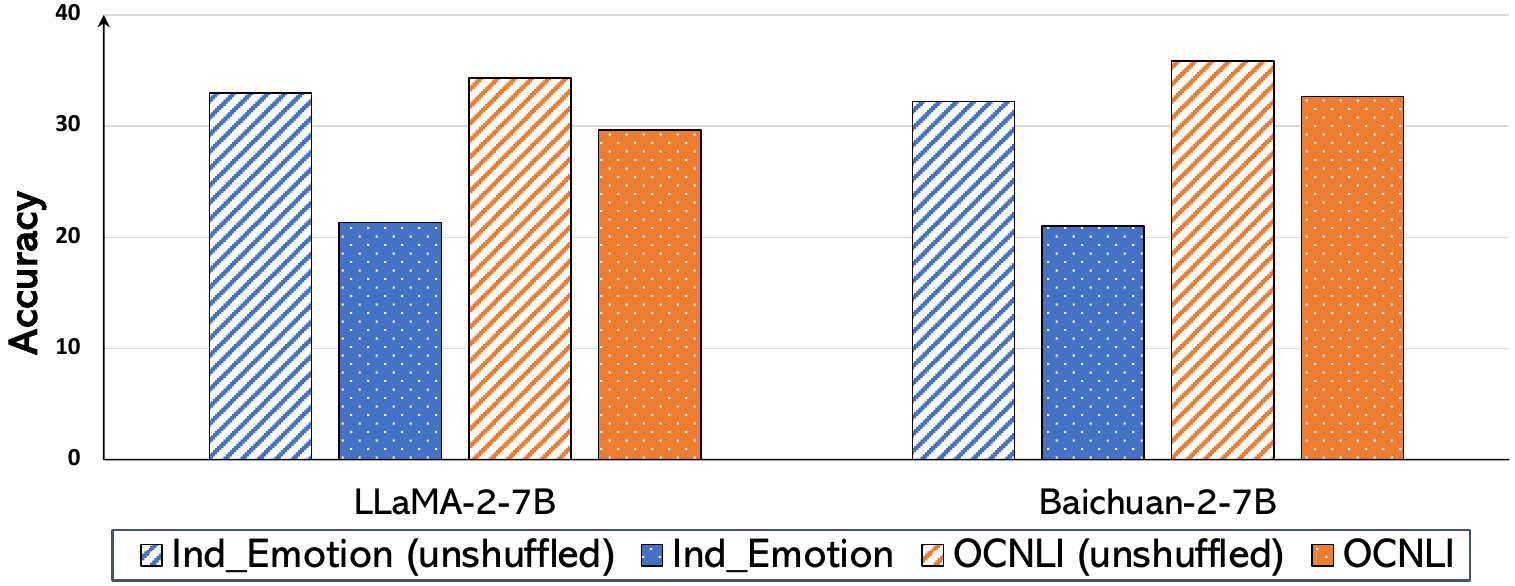}
            \caption{Effect on label order. Performance varies when labels are shuffled, revealing inherent label biases.}
            \label{fig:label_shuffling}
        \end{figure}

        \textbf{Instruction Sensitivity} Early methods for training LLMs to follow instructions primarily use task instruction sets, which are compiled by combining task instruction templates with instances from standard NLP tasks \cite{chung2022scaling}. However, such approaches often fall short of capturing the intricacies of practical user instructions, as these instructions tend to originate from artificial NLP tasks designed to test specific aspects of machine capabilities. Real-world user instructions, on the other hand, are significantly more diverse and complex \cite{ouyang2022training,wang2024resilience}, and it is necessary to evaluate the performance under varied instructions. Therefore, we build 5 human paraphrased instructions with NLP experts for each dataset. 
        We show in Figure~\ref{fig:instruction_templates} about the LLaMA-2 and ChatGPT models on their performance with five instructions and witnessed that ChatGPT models are more robust to instruction paraphrases. Some instructions possess the ability to unlock the model's full potential, potentially surpassing its more efficient counterparts, which may lead to biased evaluation~(\textit{\textbf{Our Finding 1}}). Hence, it becomes crucial to utilize multiple instructions to obtain a more comprehensive assessment of model capabilities. Evaluating the model's resilience to paraphrased instructions is also a significant aspect. To report performance, we employ the median value derived from five instructions as the ultimate result.

    \begin{figure*}[t]
        \centering
         \includegraphics[width=1.00\textwidth]{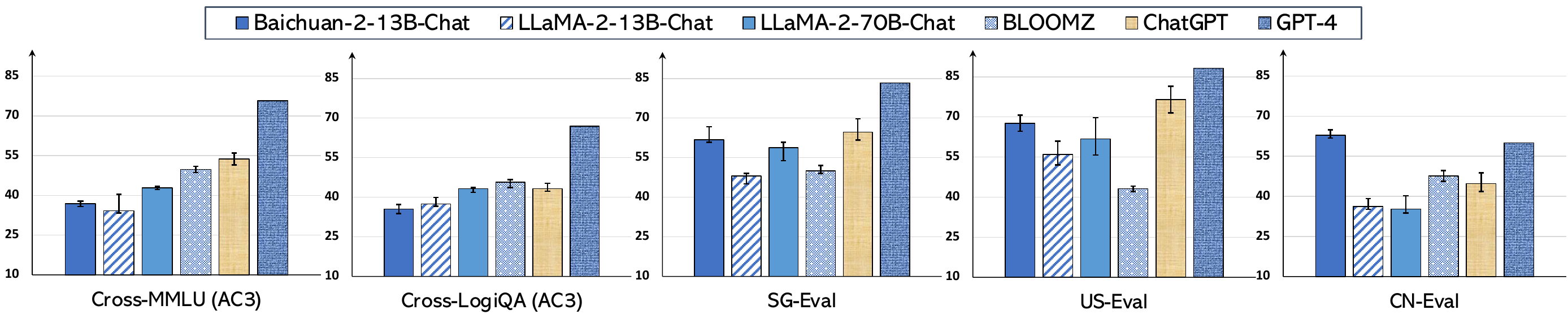}
        \caption{Evaluation results of six representative LLMs on a subset of \textit{SeaEval}. AC3 and Accuracy scores are reported. The error bar covers performances from five different instruction templates.}
        \label{fig:eval_results}
    \end{figure*}
        
        \textbf{Exposure Bias on Label Arrangements}
            Recent work demonstrates that LLMs have inherently exhibited exposure biases from many factors \cite{fei2023mitigating} including majority label bias, recency bias, and common token bias \cite{zhao2021calibrate}. In our study, we found the positional arrangement of labels is a potential source of exposure bias, especially for smaller-sized models. Figure~\ref{fig:label_shuffling} shows the results of LLaMA-2 and Baichuan-2 models on two datasets. We observe that some models are prone to rely on intrinsic biases of label arrangements when making predictions which lead to higher evaluation results. Ignoring such patterns could raise unanticipated advantages on specific models (\textit{\textbf{Our Finding 2}}). Therefore, in \textit{SeaEval}, we shuffle all labels whenever possible to avoid exposure biases on label arrangements. Note that for position-sensitive labels such as `all above', we manually keep their order unchanged.

    \begin{figure*}[t]
        \centering
         \includegraphics[width=1.00\textwidth]{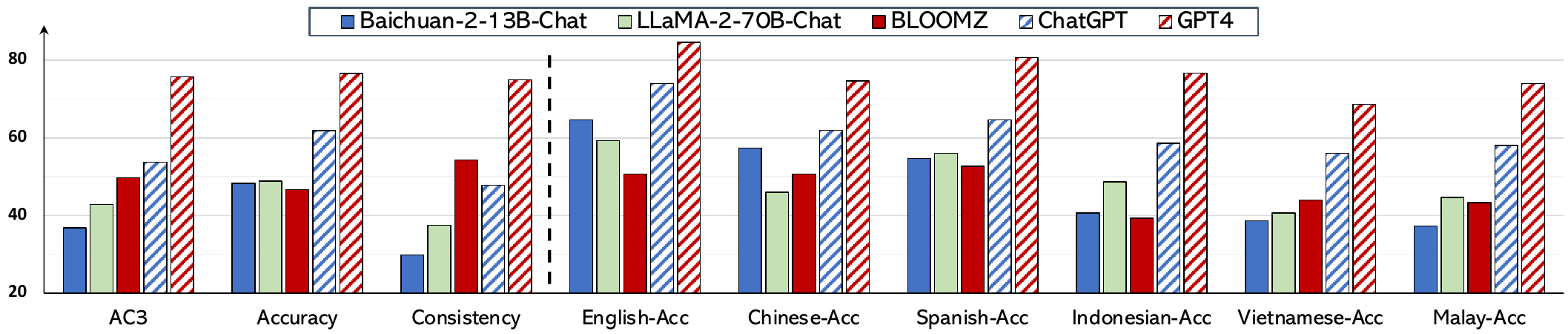}
        \caption{Evaluation results on Cross-MMLU. Both overall and language-specific scores are shown.}
        \label{fig:cross-mmlu-detailed}
    \end{figure*}

\section{Evaluation Results and Discussion}

    We show the evaluation results on five datasets for cross-lingual consistency and cultural reasoning in Figure~\ref{fig:eval_results} and our key findings are as follows.

    Firstly, GPT-4 demonstrates outstanding performance on most datasets, surpassing others by a substantial margin across cultures and languages, demonstrating its superior capability in handling multilingual tasks.
    
    Second, becoming an expert in cultural knowledge necessitates extensive pre-training with a diverse and extensive collection of multilingual textual data such as books, articles, websites, historical documents, and cultural artifacts. Baichuan-2 model has shown remarkable performance in understanding Chinese culture (CN-Eval), even outperforming GPT4. 
    In contrast, LLaMA models are primarily focused on English, with approximately 90\% of English pre-training data. This specialization makes them less proficient in handling multilingual and multicultural scenarios.

    Detailed results regarding cross-lingual consistency are presented in Figure~\ref{fig:cross-mmlu-detailed}. The report includes comprehensive evaluation metrics: \textit{AC3}, \textit{Accuracy}, \textit{Consistency}, and \textit{Accuracy} for each language. The consistency score clearly demonstrates that BLOOMZ stands out for its better performance in aligning knowledge across languages, solidifying its position as a leading open-source multilingual foundational model. Even being the worst in overall accuracy, BLOOMZ surpasses ChatGPT in cross-lingual consistency, achieving a score of 54\% compared to 47\%. However, they are still showing unsatisfactory consistency scores, highlighting the inconsistency in the sharing of common knowledge across various languages~(\textit{\textbf{Our Finding~3}}). While GPT-4 achieves a 75\% consistency score, it drops to 64\% when $s=6$ as shown in Figure~\ref{fig:cross_consis_s_mmlu}, which suggests ample opportunity to further enhance cross-lingual knowledge alignment, aiming for optimal multilingual models.

    Last, when assessing models' accuracy in individual languages, it is evident that their problem-solving capability in English usually surpasses that in other selected languages. This illustrates that the proficiency of models varies unevenly across different languages~(\textit{\textbf{Our Finding 4}}). Compared to high-resource languages, the performance of low-resource languages is inferior. For example, English, Chinese and Spanish rank within the top 5 out of 46 languages present in BLOOMZ corpus. Therefore, the multilingual foundation model's capability in low-resource languages needs to be further improved to match the more centralized languages. The disparity in cross-lingual consistency highlights the need for more robust alignment efforts. Particularly under low-resource constraints, enhancements in this aspect have the potential to elevate the overall performance across all languages through effective knowledge transfer, facilitating further development of multilingual language models~\cite{kulshreshtha-etal-2020-cross,huang2023not,zhu2023extrapolating,muennighoff-etal-2023-crosslingual}.

\section{Conclusions}

    We introduced \textit{SeaEval} benchmark for multilingual foundation model evaluation, grounded in comprehensive experimentation across languages, models, tasks, and datasets. \textit{SeaEval} encompasses 29 datasets, including 7 new ones for cultural understanding and cross-lingual consistency. Our empirical analysis demonstrates four key findings on the capabilities of multilingual foundation models: 1) Sensitivity to paraphrased instructions; 2) Exposure bias of label arrangements, 3) Inconsistent performance across multilingual questions that are semantically equivalent, and 4) Imbalanced multilingual proficiency. These findings accentuate the importance of more generalizable semantic representations and enhanced multilingual contextualization. We hope that our endeavors in \textit{SeaEval} can pave the way for more in-depth investigations into multilingual and multicultural tasks using foundation models.

\section*{Limitations}
   
    In this study, our primary focus is multilingual foundation models' language capabilities. Nonetheless, there remain several evaluation aspects to be included in order to provide a complete reflection of the capability of multilingual foundation models in practical applications.
    
    First, there is a need for the inclusion of more languages and cultural reasoning datasets. Expanding the linguistic and cultural diversity within the benchmark is a resource-intensive endeavor, as the acquisition of suitable datasets for various languages and culture-related contexts can be challenging. Nevertheless, as we aspire for this benchmark to comprehensively cover a wide range of languages, there is a pressing need to explore automated methods for data collection. Such an approach can help ensure the acquisition of high-quality datasets while mitigating the resource-intensive nature of manual data curation, thereby enhancing scalability.
    
    Second, \textit{SeaEval} ensures a robust quantitative evaluation benchmark, incorporating datasets that facilitate more straightforward performance quantization. In real-world usage cases, foundation models are also used for information-seeking purposes, where users may pose subjective questions and engage in dialogues. This poses challenges in evaluating the faithfulness, expertise and engagement during interactions. Existing approaches adopt powerful FMs as the evaluation criteria which may not necessarily replicate the judgments from humans~\cite{zheng2023judging}, underscoring the necessity of practical automatic assessment approaches for open-ended questions.

    Third, but certainly not the least, safety and efficiency are two important dimensions of FMs. Ensuring the safety of models in real-time and dynamic contexts is critical, especially to avoid generating harmful or biased content. Meantime, striking a balance between the effectiveness and efficiency of FMs is challenging and requires more ongoing research efforts. Therefore, our pursuit of a comprehensive benchmark should extend to these vital dimensions of model performances.

\section*{Acknowledgement}

    
    This work is supported by the National Research Foundation, Singapore under its AI Singapore Programme (AISG Award No: AISG2-GC-2022-005). The computational work for this article was partially performed on resources of the National Supercomputing Centre (NSCC), Singapore. It is also partially supported by Cloud TPUs from Google’s TPU Research Cloud (TRC). We thank Xunlong Zou and Geyu Lin for participating in research discussions, and Siti Umairah Md Salleh, Siti Maryam Binte Ahmad Subaidi, Nabilah Binte Md Johan, Wiwik Karlina, Xuan Long Do, Fabian Ritter Gutierrez and Ayrton San Joaquin for their contribution to cross-lingual resource construction and verification.

\bibliography{anthology,custom}

\appendix

\section{Related Work}

    \subsection{Existing Benchmarks}
        
        The field of LLM evaluation is expanding quickly owing to the rapid development of model capabilities. \citet{chang2023survey} provides a comprehensive review of different evaluation methods. Even though there are thousands of languages around the world, the vast majority of LLM evaluation benchmarks concentrate on English or Chinese~\cite{liang2022holistic,alpaca_eval,zhong2023agieval}, which has a solid foundation of well-annotated resources. They are mainly focused on complex reasoning datasets which are normally collected from human examinations including SAT, math tests or Chinese college examinations~\cite{zhong2023agieval,hendryckstest2021,huang2023ceval,li2023cmmlu}. \citet{gu2023xiezhi} gathered questions from various disciplines like economics, jurisprudence and literature. Besides the subjective test, \citet{alpaca_eval} and \citet{bai2023benchmarking} propose to use LLMs as the judger to provide objective evaluations on generated content for objective scores and implement pairwise model ranking.

        There are pioneering efforts on multilingual large language model evaluation. \citet{lai2023chatgpt} propose to evaluate large language models for their multilingual capability with a series of classic NLP tasks. \citet{zhang2023m3exam} expands multilingual evaluation to 9 languages associated with a toolkit. \citet{zhu2023multilingual} evaluate LLM with machine translation test sets which are multilingual inherently. In this work, we expand multilingual foundation model evaluation benchmarks beyond combinations of monolingual tasks.

    \begin{figure*}[t]
        \centering
         \includegraphics[width=1.00\textwidth]{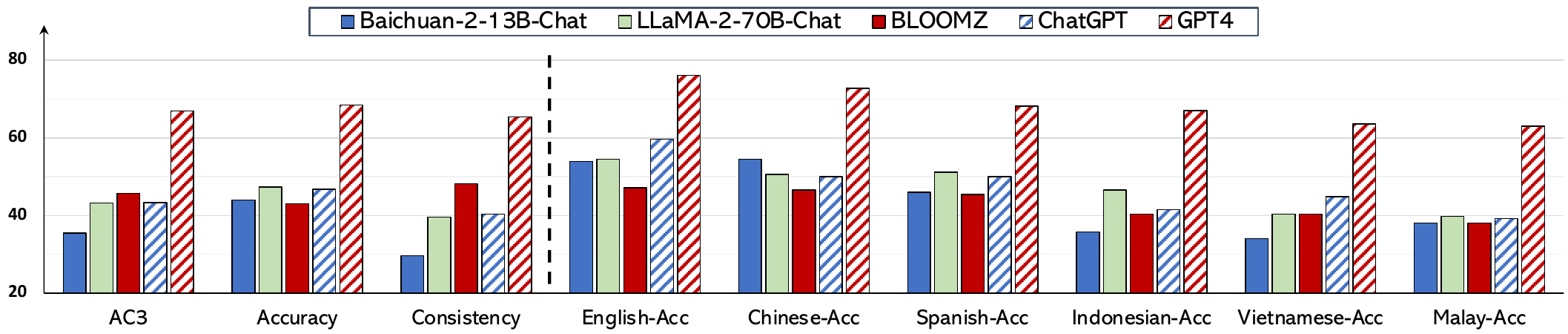}
        \caption{Evaluation results on Cross-LogiQA. Both overall and language-specific scores are shown. }
        \label{fig:cross-logiqa-detailed}
    \end{figure*}

    \begin{table}[t]
        \centering
        \begin{adjustbox}{width=0.5\textwidth,center}
        \begin{tabular}{ l | l }
        \toprule
         \textbf{Question} & \emph{Which of the following items would be considered } \\ 
         & \emph{the least suitable gift to bring to a Singaporean} \\ 
         & \emph{family during Lunar New Year?} \\ 
         & \emph{(A) An pineapple} \\
         & \emph{(B) Cash money} \\
         & \emph{(C) Two organges} \\
         & \emph{(D) A red packet} \\
         \midrule
         \textbf{Multicultural} & \textcolor{alizarin}{\textbf{Multilingual Understanding}} \\
         \textbf{Reasoning Steps} & $\cdot$ Pineapple = Ong lai (Hokkien), sounds\\ 
         & like `good fortune to come' \\
         
         & $\cdot$ Orange sounds like `good luck' in Cantonese \\
         & \textcolor{alizarin}{\textbf{Cultural Preferences}} \\
         & $\cdot$ Chinese like `double' as a representation \\
         & of unity, completeness and harmony. \\
         & $\cdot$ Red packet is preferred as associated with luck, \\
         & prosperity and happiness.\\
        \midrule
         \textbf{Answer} & (B) Cash Money \\
    
        \bottomrule
        \end{tabular}
        \end{adjustbox}
        \caption{
            The 2nd example of cultural reasoning from SG-Eval.
        }
        \label{tab:sample_sg_eval_example2}
    \end{table}

    \subsection{Foundation Language Models}
        The foundation language models, as general task solvers, include both pre-trained language models and their instruction-tuned variants. ChatGPT~\cite{ouyang2022training} and GPT4~\cite{openai2023gpt} are showing superior capabilities across various applications. A series of foundation models are released afterwards including Flan-T5~\cite{chung2022scaling}, Alpaca~\cite{alpaca}, Vicuna~\cite{vicuna2023} and LLaMA-2~\cite{touvron2023llama}. Their multilingual capability is inferior to English due to the unbalanced training corpus and vocabulary settings. For applicable to multilingual scenarios, a series of bilingual or multilingual models are proposed by pertaining from scratch (XGLM~\cite{lin-etal-2022-shot}, BLOOM~\cite{scao2022bloom}, ChatGLM~\cite{du2022glm}), expansion of vocabulary sizes~\cite{cui2023efficient} or aligning multilingual instructions~\cite{zhu2023extrapolating}. In the foreseeable future, we anticipate a surge of multilingual language models, underscoring the need for effective multilingual LLM evaluation benchmarks.

\section{Selected Models}

    In this work, we evaluate the performance of various large language models on our benchmark datasets. They show disparate capabilities in various tasks. The included models are Flan-T5 (Flan-T5-Small, Flan-T5-Base, Flan-T5-Large, Flan-T5-XL, FLAN-T5-XXL, FLAN-UL2)~\cite{chung2022scaling}, LLaMA-1 (LLaMA-7B, LLaMA-13B, LLaMA-30B, LLaMA-65B)~\cite{touvron2023llama-1}, LLaMA-2 (LLaMA-2-7B, LLaMA-2-7B-Chat, LLaMA-2-13B, LLaMA-2-13B-Chat, LLaMA-2-70B, LLaMA-2-70B-Chat)~\cite{touvron2023llama}, Baichuan (Baichuan-7B, Baichuan-13B, Baichuan-13B-Chat, Baichuan-2-7B, Baichuan-2-7B-Chat, Baichuan-2-13B, Baichuan-2-13B-Chat)~\cite{yang2023baichuan}, Alpaca-7B~\cite{alpaca}, Vicuna (Vicuna-7B-v1.3, Vicuna-13B-v1.3, Vicuna-7B-v1.5, Vicuna-13B-v1.5, Vicuna-33B-v1.3)~\cite{vicuna2023}, ChatGLM (ChatGLM-6B, ChatGLM2-6B)~\cite{du2022glm}, BLOOM (BLOOMZ-7B1, MT0-XXL)~\cite{scao2022bloom}, Colossal-LLaMA-2-7B-Base, ChatGPT (gpt-3.5-turbo-0613, gpt-4-0613)~\cite{openai2023gpt}. In this paper, we report the result of the following models as a representative set considering their overall performance and multilingual support.

    \begin{itemize}
        \item \textbf{Baichuan-2}: is an open-source multilingual language model with emphasis on English and Chinese. It shows competitive performance compared to models of the same size and generally outperforms LLaMA-2 model through better pre-training and human alignment techniques. Baichuan-2-13B-Chat is selected in our experiments.
        \item \textbf{LLaMA-2}: is an open-source language model released by Meta. Even though it supports multilingual, LLaMA is trained with most data (close to 90\%) in English which makes it an English-centric model. It performs the best for English use cases than other languages. The Chat variant is further tuned for improved helpfulness and safety. LLaMA-2-13B-Chat and LLaMA-2-70B-Chat are selected in our experiments.
        \item \textbf{BLOOMZ}: is the leading open-source multilingual large language model further tuned with diverse instructions. It supports over 40 languages with a more balanced pertaining and fine-tuning corpus. Note that BLOOMZ is instruction-tuned with supervised datasets which may cause supervision leakage on certain datasets (e.g. SAMSum, DREAM) and unjustified comparison. BLOOMZ-7B1 is selected in our experiments.
        \item \textbf{ChatGPT}: is closed-source model developed by OpenAI. It has good multilingual support and demonstrates more robust performance compared to open-source models. The model is updating over time and we select GPT3.5 (referred to as ChatGPT) and GPT4 on version 0613 in all our experiments.
    \end{itemize}

    \begin{figure}[ht]
    \centering
    \includegraphics[width=0.45\textwidth]{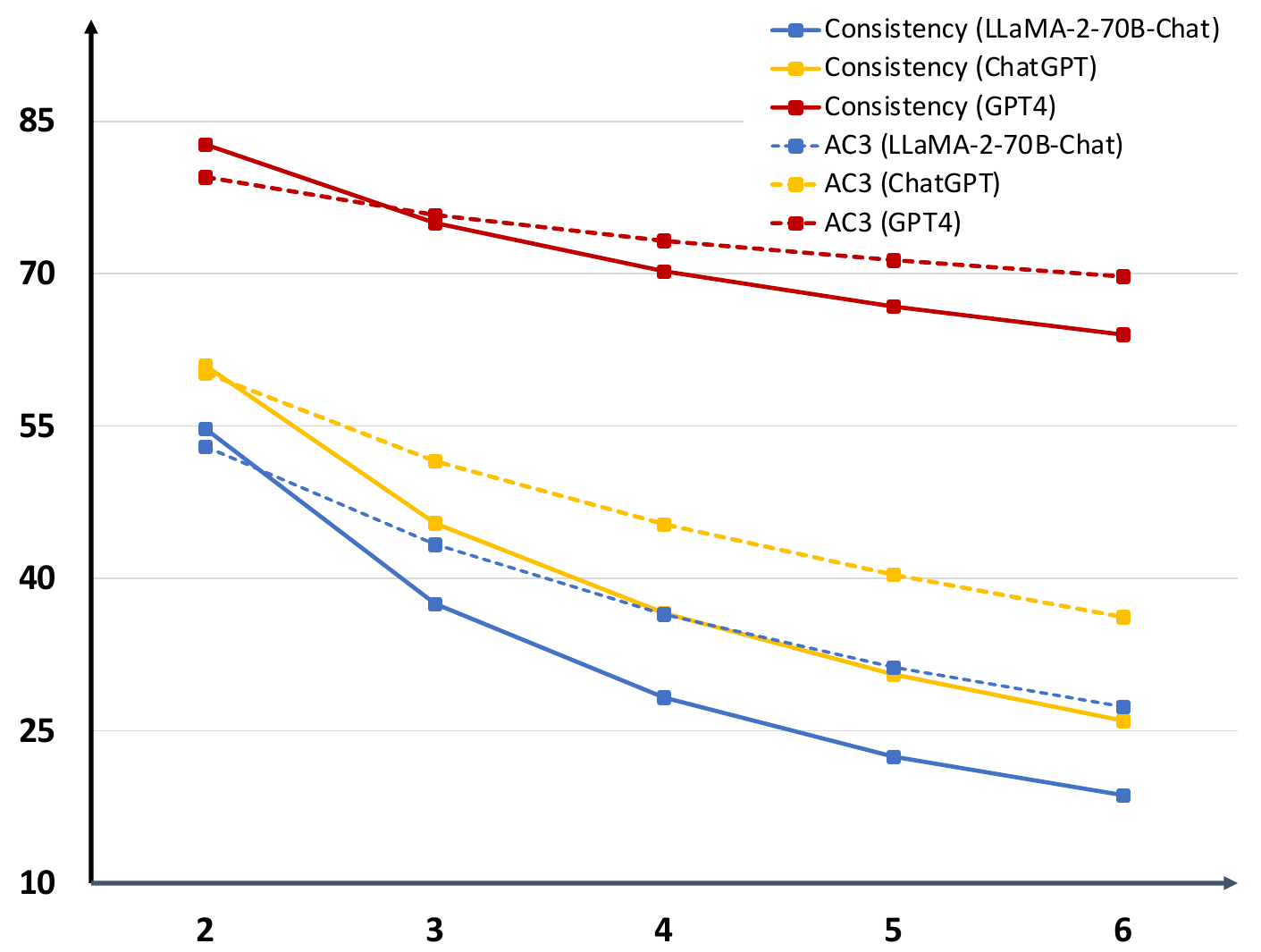}
    \caption{Consistency and AC3 Scores with $s\in[2,6]$ on Cross-MMLU dataset. (Filipino excluded)}
    \label{fig:cross_consis_s_mmlu}
    \end{figure}

    \begin{figure}[ht]
    \centering
    \includegraphics[width=0.45\textwidth]{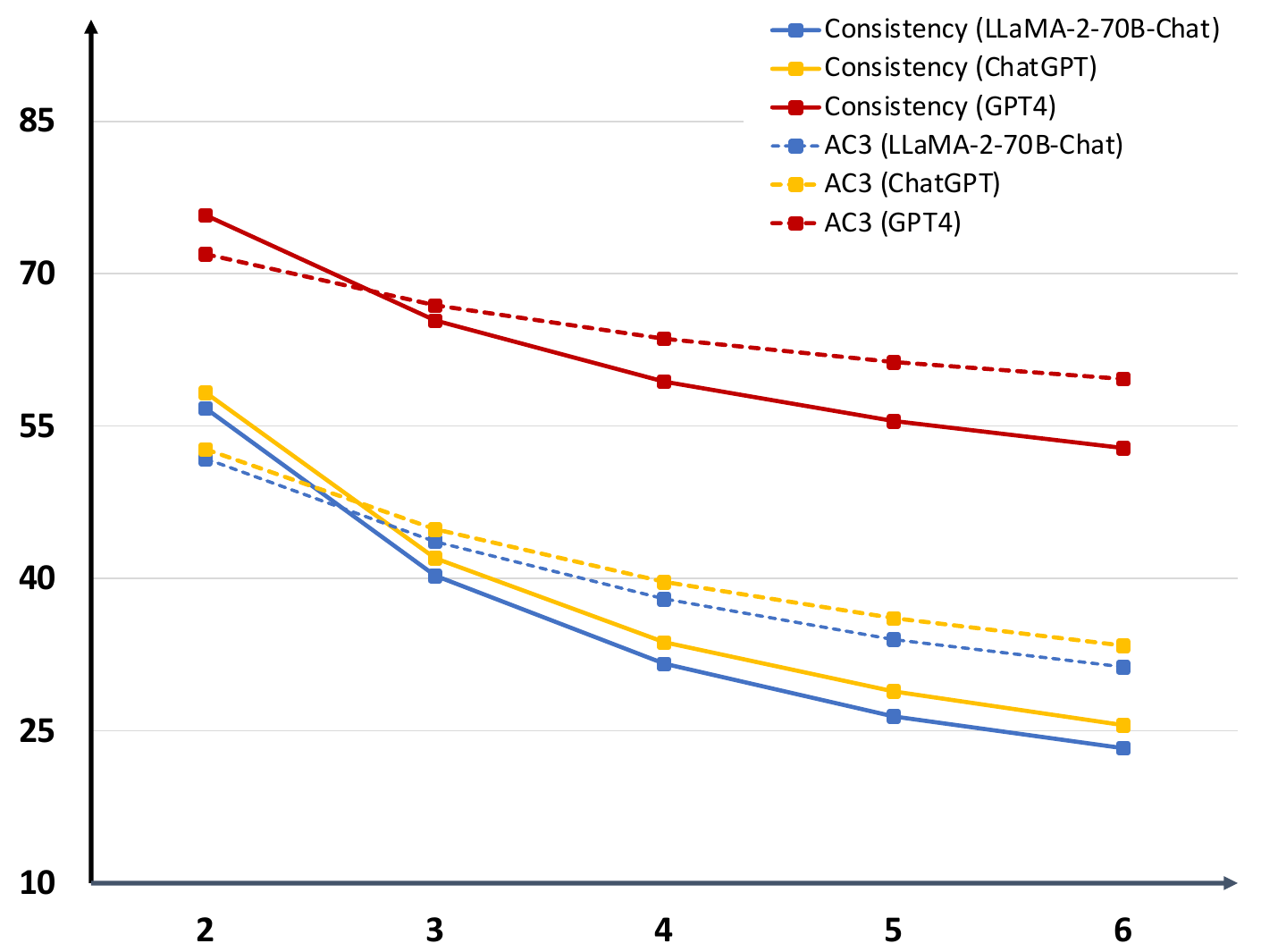}
    \caption{Consistency and AC3 Scores with $s\in[2,6]$ on Cross-LogiQA dataset. (Filipino excluded)}
    \label{fig:cross_consis_s_logiqa}
    \end{figure}

\section{Cross-Lingual Consistency}

    In this paper, we spot the cross-lingual inconsistency problem for multilingual foundation models. To better evaluate this aspect, we collect and propose two new datasets with respective metrics. In this section, we provide more analysis on the cross-lingual consistency study and the effectiveness of $s$ in the proposed \emph{Consitency} and \emph{AC3} metrics. 

    The evaluation results for Cross-LogiQA dataset is depicted in Figure~\ref{fig:cross-logiqa-detailed}. As the leading open-source multilingual model, BLOOMZ outperforms other open-source models and ChatGPT in terms of consistency but falls short in achieving high performance in specific languages. This suggests that BLOOMZ provides more consistent answers across languages, possibly due to its training on a more balanced multilingual corpus and fine-tuning with multilingual instruction data to improve cross-language alignment. As observed in Figure~\ref{fig:cross-mmlu-detailed}, the performance is better for higher-resource languages such as English, Chinese, and Spanish compared to lower-resource languages like Indonesian, Vietnamese, and Malay. GPT4 surpasses other models in both "Accuracy" and "Consistency," highlighting significant potential for further enhancement of all other models.

    In Section~\ref{sec:evaluation_protocols}, AC3 score is presented, taking into account both accuracy and consistency scores. We have one tolerance hyperparameter $s$ which requires the answers to be consistent across $s$ languages to be rewarded in consistency score. We deploy $s=3$ as the default hyperparameter in above experiments. Here, we conduct more systematic study of $s$ with its effect on the final scores.

    The results on Cross-MMLU and Cross-LogiQA are shown in Figure~\ref{fig:cross_consis_s_mmlu} and \ref{fig:cross_consis_s_logiqa}. As $s$ increases, the consistency score has dropped dramatically for all models. Among all three models, the performance of GPT4 drops the least, indicating a robust consistency alignment across languages. Even for ChatGPT, when $s=6$, the consistency score downgrades to around 26\% on both Cross-MMLU and Cross-LogiQA datasets. It indicates that only 26\% of the cases that ChatGPT is selecting the same answer for the same question across six languages. Hence, we opt for $s=3$ as the default value for two primary reasons: 1) to facilitate evaluation across multilingual language models, even when not all 6 languages are supported, and 2) to allow for a certain level of tolerance regarding language consistency, without imposing the strict requirement of complete consistency among all responses. This approach can be seen as a more lenient method of assessing language consistency. In the future, as the model's multilingual capabilities continue to advance, we can increase $s$ accordingly.

\section{Evaluation Protocols}

    When assessing foundation models, two distinct settings come into play: zero-shot and few-shot. In this study, we predominantly rely on zero-shot evaluation as the chosen method for all models, primarily for two compelling reasons. Firstly, zero-shot evaluation aligns more closely with real-world application scenarios, where users interact directly with deployed models without undergoing explicit training. Secondly, it's worth noting that even without the process of fine-tuning using human instructions, these models display a noteworthy ability to comprehend and adhere to emerging instructions. Besides, it brings additional benefits to avoid uncertainty caused by the in-context few-shot samples and potential exposure biases.

    Another challenge in evaluation is the unstructured form of outputs from large language models. Unlike previous discriminative models, language generation models produce the answer represented by free text. Therefore, there is a gap between the generated content and the ground-truth answers, especially for multi-choice questions. Therefore, we develop a heuristic algorithm to decide the mapping. In general, we first split the answer into sentences. For each sentence, we detect whether the choice symbols (e.g. (A), (B), (C), (D)) exist. If none of them exists, we detect whether the choice description exists. If exists, we count the sentence as the symbol of such label and N.A. otherwise. Finally, we perform majority voting from all sentences as the final answer. From the experiments, we found the algorithm is robust enough to link the generated content with labels among diverse models and languages. Therefore, we chose this algorithm after careful comparison with a few other variants.

\section{Full Experimental Results and Analysis}

\begin{figure*}[t]
        \centering
         \includegraphics[width=1.0\textwidth]{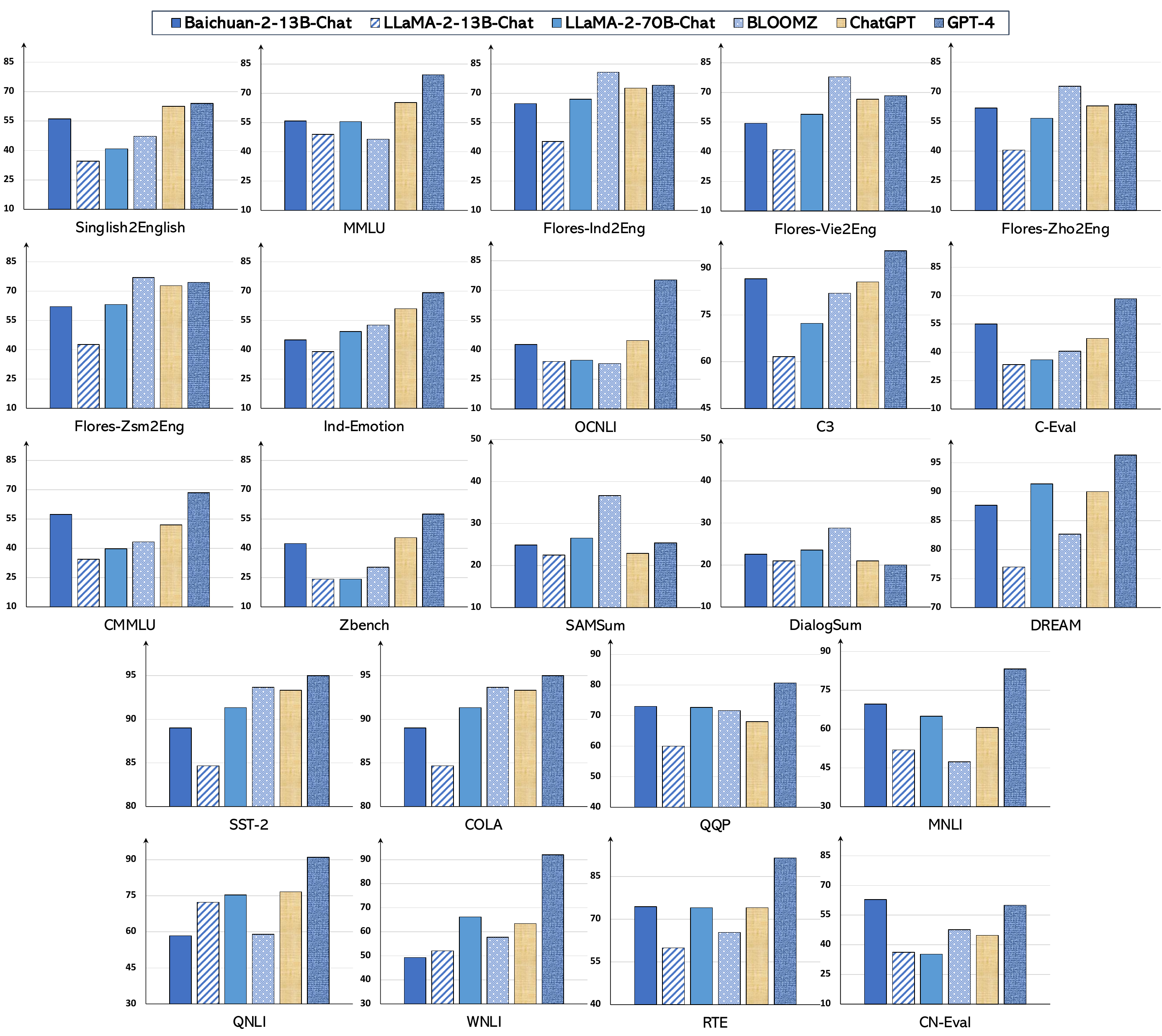}
        \caption{Evaluation results of LLMs on the rest of SeaEval tasks as supplement of Figure~\ref{fig:eval_results}.}
        \label{appendix:fig:eval_results_1}
    \end{figure*}

    Besides the evaluation on 5 datasets shown in Figure~\ref{fig:eval_results}, the full evaluation results on the other 23 datasets are shown in Figure~\ref{appendix:fig:eval_results_1}. From the results, we spot several key findings.

    First, Baichuan-2-13B-Chat surpasses LLaMA-2-13B-Chat not only in Chinese tasks but also in tasks in English and other languages. Despite being pre-trained with 2T tokens mostly in English, LLaMA-2 falls short in terms of its English proficiency. This highlights the critical role of data quality and diversity. Effective data collection and post-processing play a pivotal role in the development of large language models.

    Secondly, when it comes to a multilingual context, BLOOMZ exhibits lower competitiveness in comparison to other models. Despite being trained on a dataset that incorporates more than 40 languages and pre-trained using a more evenly balanced corpus from these languages, it fails to showcase superior performance in various multilingual tasks. This could be attributed to ineffective pre-training and the limitations imposed by the model's size, which consists of 7 billion parameters. It is worth noting that BLOOMZ does display exceptional performance in specific datasets, such as SAMSum and Flores. This can be attributed to the direct fine-tuning of the model with supervision, making it inappropriate to draw direct comparisons with datasets as outlined in~\citep{muennighoff-etal-2023-crosslingual}. Nevertheless, the cross-lingual consistency of the BLOOMZ model is good due to its cross-lingual generalization through multitask finetuning~\cite{muennighoff-etal-2023-crosslingual}.
    
    Lastly, GPT4 surpass ChatGPT in various aspects including multilingual capability. However, due to its commercial nature, it is generally hard to conduct transparent research with such models.

\section{More Examples}
\label{appendix:more_examples}

    To have a direct interpretation of the newly proposed six datasets, we further illustrate samples and instructions with English translations in Table~\ref{tab:seaeval_examples}. Examples of SG culture questions are shown in Table ~\ref{tab:sample_sg_eval} and~\ref{tab:sample_sg_eval_example2}.

\begin{figure*}[ht]
    \centering
    \includegraphics[width=0.8\textwidth]{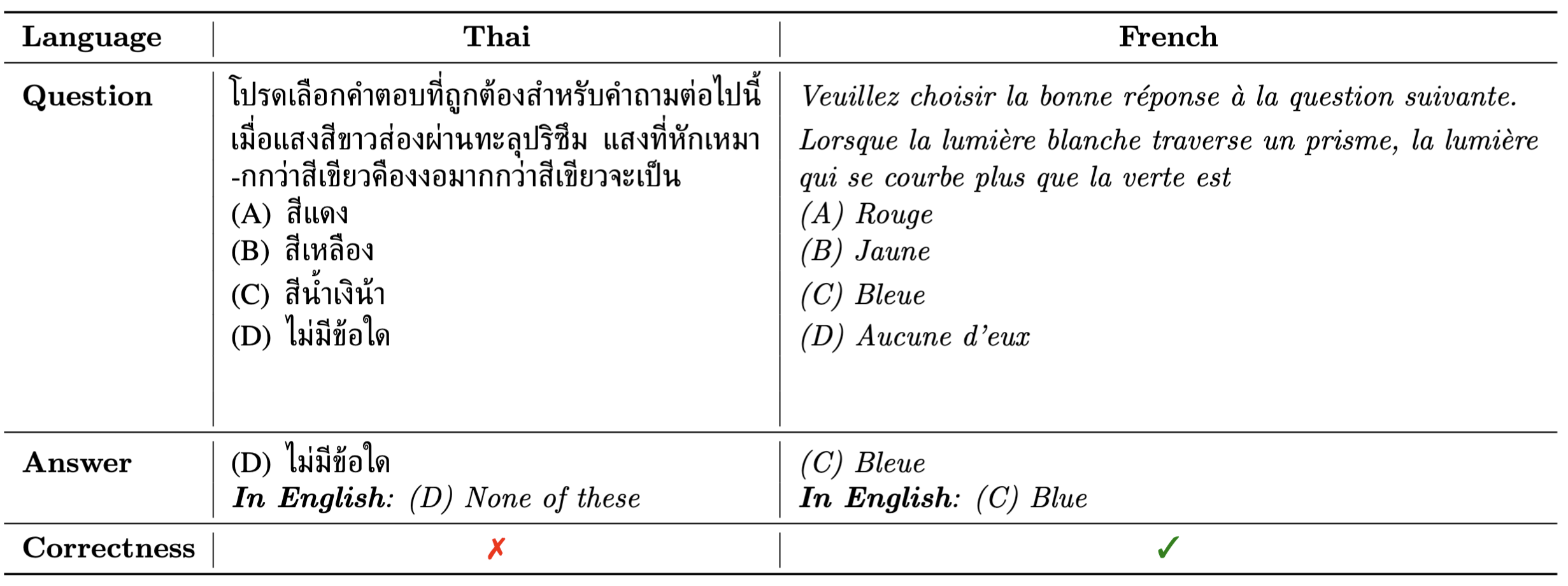}
    \caption{An example from our Cross-MMLU dataset on Thai and French}
    \label{fig:cross_mmlu_example_2}
    \end{figure*}

\begin{figure*}[ht]
    \centering
    \includegraphics[width=0.7\textwidth]{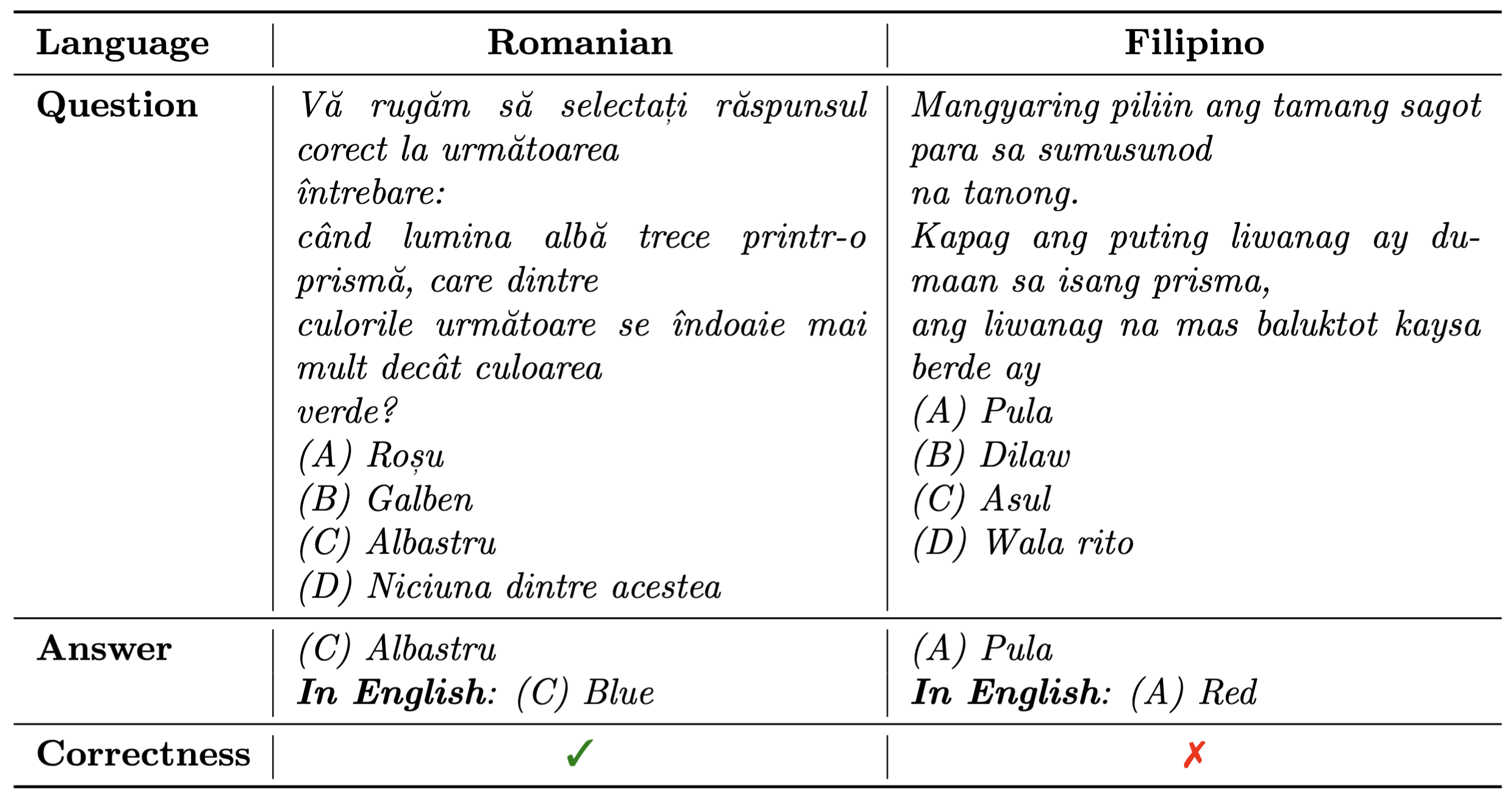}
    \caption{An example from our Cross-MMLU dataset on Romanian and Filipino}
    \label{fig:cross_mmlu_example_3}
    \end{figure*}

\begin{figure*}[ht]
    \centering
    \includegraphics[width=0.9\textwidth]{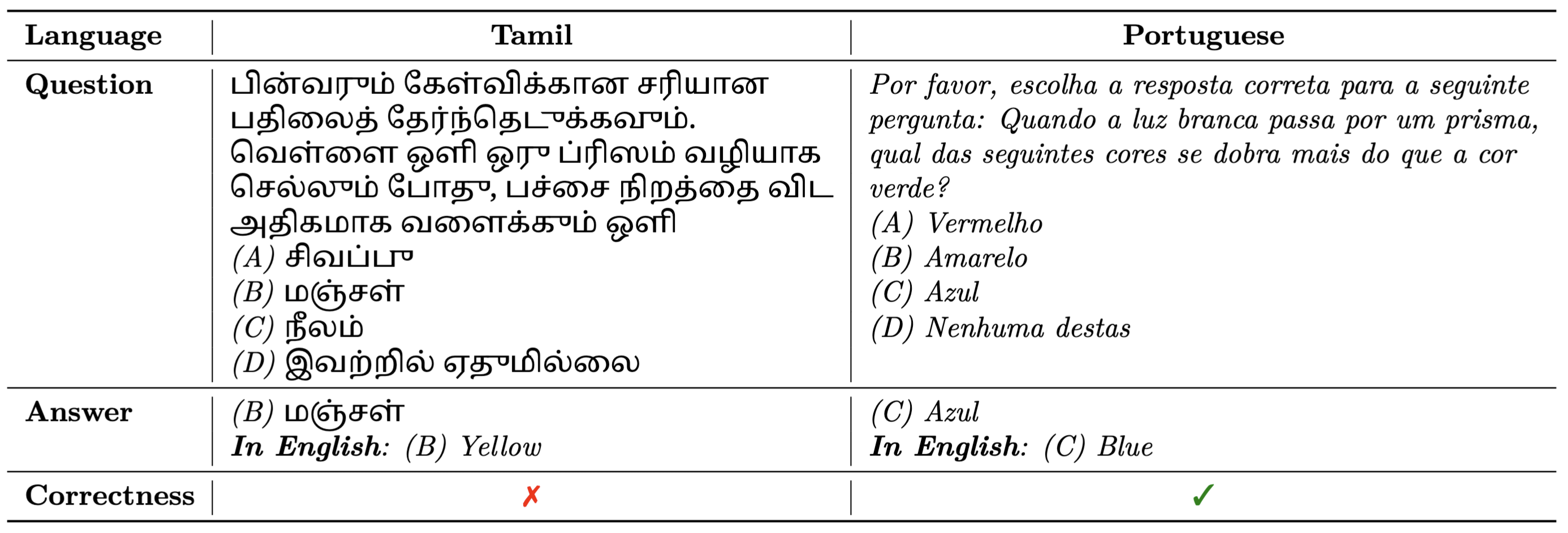}
    \caption{An example from our Cross-MMLU dataset on Tamil and Portuguese}
    \label{fig:cross_mmlu_example_5}
    \end{figure*}

\begin{figure*}[ht]
    \centering
    \includegraphics[width=0.45\textwidth]{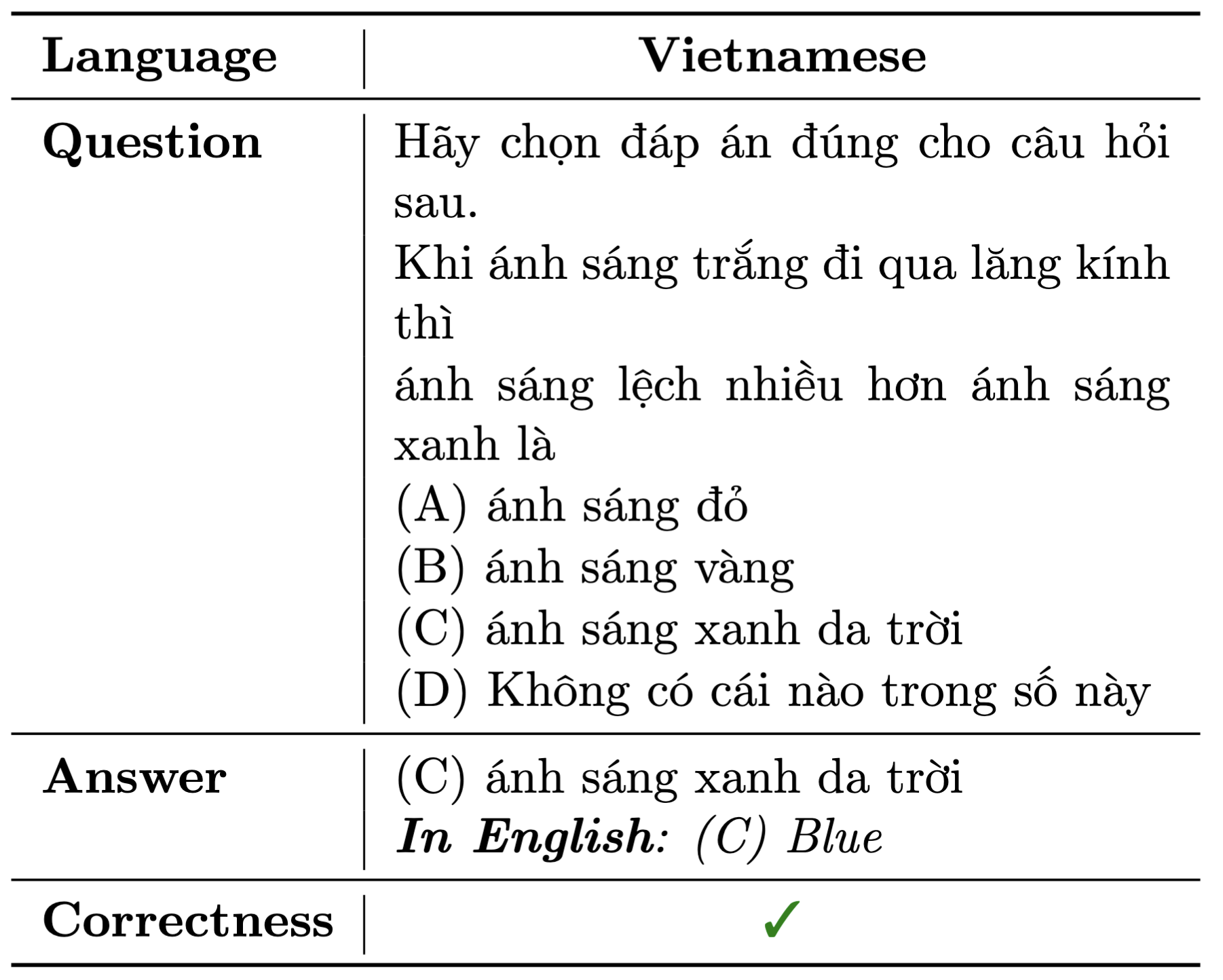}
    \caption{An example from our Cross-MMLU dataset on Vietnamese}
    \label{fig:cross_mmlu_example_4}
    \end{figure*}

\begin{figure*}[ht]
    \centering
    \includegraphics[width=0.45\textwidth]{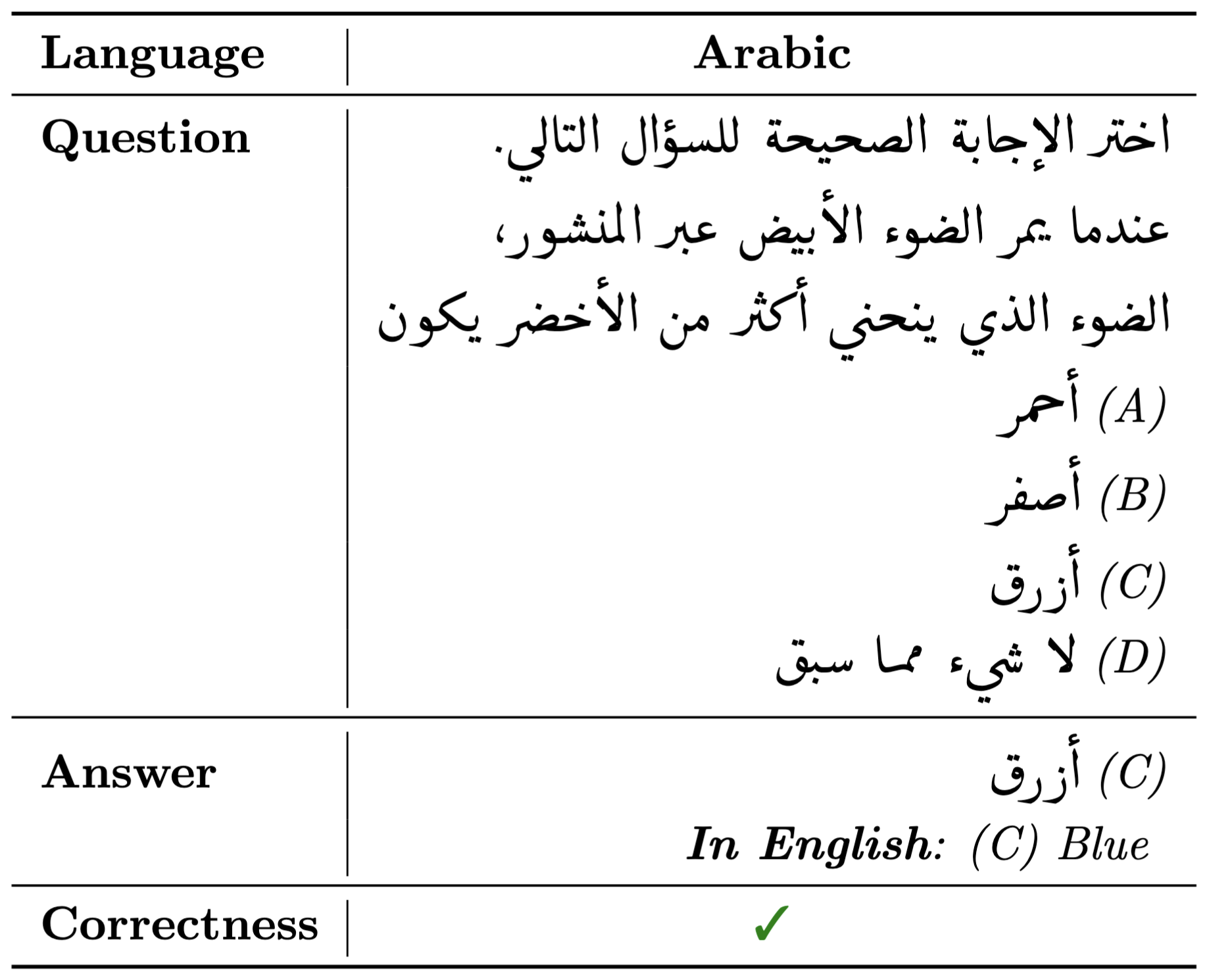}
    \caption{An example from our Cross-MMLU dataset on Arabic}
    \label{fig:cross_mmlu_example_6}
    \end{figure*}

    For evaluating cross-lingual consistency, we introduce two datasets: Cross-MMLU and Cross-LogiQA, featuring parallel questions in 6 different languages. In this section, we delve deeper into cross-lingual inconsistency phenomena across a broad range of languages. To achieve this, we expand our sample to include questions in 16 languages and prompt ChatGPT for answers. The outcomes are detailed in Table~\ref{tab:cross-lingual-alignment-example},~\ref{tab:cross-lingual-alignment-example-malay-korean} and Figure~\ref{fig:cross_mmlu_example_2},~\ref{fig:cross_mmlu_example_3},~\ref{fig:cross_mmlu_example_5},~\ref{fig:cross_mmlu_example_4},~\ref{fig:cross_mmlu_example_6}. The languages encompassed in this study are English, Chinese, Indonesian, Spanish, Thai, French, Korean, Malay, Turkish, German, Romanian, Filipino, Tamil, Portuguese, Vietnamese and Arabic. The results substantiate the existence of cross-lingual consistency issues across different languages, underscoring the need for increased attention to this matter.

\section{Annotators}

    The annotators consist of both full-time employees and PhD students. Full-time employees did not receive additional compensation for their annotation work but considered it as part of their regular working hours. Conversely, PhD students had their annotation time recorded and were compensated with fixed-hour claim rates. On average, one round of correction for each language on each dataset took approximately 3-5 hours, varying depending on the languages involved.

    \begin{table*}[t]
        \centering
        \begin{adjustbox}{width=0.90\textwidth,center}
        \begin{tabular}{ l | l | l }
        \toprule
                
        \multicolumn{3}{l}
        {\sethlcolor{hlgreen}\hl{\textbf{Multicultural and Multilingual Understanding}}}
        \\ \hline \hline
        
        \multirow{7}{*}{\textbf{SG-Eval}} & Instruction & \emph{Please carefully read the following question and select the most appropriate answer from the choices.}  \\ 

        \cline{2-3}
        
        & \multirow{6}{*}{Sample} & \emph{Which drink in Singapore has the highest calories?}  \\ 
        & & \emph{(A) Teh O} \\
        & & \emph{(B) Teh Siew Dai} \\
        & & \emph{(C) Kopi} \\
        & & \emph{(D) Kopi C} \\
        & & \emph{Answer: (C) Kopi} \\
        
        \hline

        \multirow{7}{*}{\textbf{US-Eval}} & Instruction & \emph{Read the following question carefully and select the correct answer from the choices.} \\

        \cline{2-3}

        & \multirow{6}{*}{Sample} & \emph{When daylight-saving time arrives in the spring how do most Americans turn their clocks?} \\
        & & \emph{(A) one hour forward} \\
        & & \emph{(B) one hour backward} \\
        & & \emph{(C) two hours forward} \\
        & & \emph{(D) two hours backward} \\
        & & \emph{Answer: (A) one hour forward} \\

        \hline

        \multirow{15}{*}{\textbf{CN-Eval}} & Instruction & \begin{CJK*}{UTF8}{gbsn} 请仔细阅读以下问题，并从选项中选择最合适的答案。 \end{CJK*} \\

        \cline{2-3}

        & \multirow{14}{*}{Sample} & \begin{CJK*}{UTF8}{gbsn} 清代官场饮茶有着特殊的程序和含义，有别于一般的茶道，主人若端茶，对客人说“请喝茶”，这表明 \end{CJK*} \\
        & &  \begin{CJK*}{UTF8}{gbsn} (A)对客人不满 \end{CJK*} \\
        & &  \begin{CJK*}{UTF8}{gbsn} (B)请客人品茶\end{CJK*} \\
        & &  \begin{CJK*}{UTF8}{gbsn} (C)对客人的尊敬\end{CJK*} \\
        & &  \begin{CJK*}{UTF8}{gbsn} (D)会谈结束送客\end{CJK*} \\
        & & \begin{CJK*}{UTF8}{gbsn} 答案: (D)会谈结束送客\end{CJK*} \\
        & & \textbf{Translation:} \\
        & & \emph{Tea drinking in officialdom in the Qing Dynasty had special procedures and meanings, which were different from}\\
        && \emph{ordinary tea ceremonies. If the host served tea and said "Please drink tea" to the guests, this meant} \\
        && \emph{(A) Dissatisfied with the guest} \\
        && \emph{(B) Invite guest to taste tea} \\
        && \emph{(C) Show Respect for guest} \\
        && \emph{(D) End the meeting and seeing off the guest} \\
        && \emph{Answer: (D) End the meeting and see off the guest} \\

        \hline

        \multirow{5}{*}{\textbf{Singlish2English}} & Instruction & \emph{Translate the following sentence from Singlish to English. Please only output the translated sentence.} \\

        \cline{2-3}

        & \multirow{4}{*}{Sample} & \textbf{Source in Singlish:} \\ 
        & & \emph{Wah this one damn shiok and underrated. The maggi goreng also damn sedap. Bro you got refined taste} \\
        & & \textbf{Target in Standard English:}\\
        & & \emph{Wow, this is super enjoyable and underrated. The Maggi Goreng is damn delicious. Brother, you have got a refined taste.} \\
        \hline\hline

        \multicolumn{3}{l}
        {\sethlcolor{pink}\hl{\textbf{Cross-Lingual Consistency}}}
        \\ \hline \hline

        \multirow{2}{*}{\textbf{Cross-MMLU}} & Instruction & \emph{Respond to the question by selecting the most appropriate answer.} \\

        \cline{2-3}

        & Sample & Shown in Table~\ref{tab:cross-lingual-alignment-example}. \\

        \hline

        \multirow{27}{*}{\textbf{Cross-LogiQA}} & Instruction & \emph{Kindly choose the correct answer from the options provided for the multiple-choice question.} \\

        \cline{2-3}

        & \multirow{26}{*}{Sample} & \textbf{English Version:} \\
        & & \emph{Content: At a gathering at which bankers, athletes, and lawyers are present, all of the bankers are athletes} \\
        & & \emph{and none of the lawyers are bankers.} \\
        & & \emph{Question: If the statements above are true, which one of the following statements must also be true?} \\
        & & \emph{(A) Some of the lawyers are not athletes.} \\
        & & \emph{(B) Some of the athletes are not lawyers.} \\
        & & \emph{(C) None of the lawyers are athletes.} \\
        & & \emph{(D) All of the athletes are bankers.} \\
        && \emph{Answer: (B) Some of the athletes are not lawyers.}\\

        & & \textbf{Chinese Version:} \\
        & & \begin{CJK*}{UTF8}{gbsn} 在银行家，运动员和律师的聚会上，所有银行家都是运动员，没有一个律师是银行家。 \end{CJK*} \\
        &&   \begin{CJK*}{UTF8}{gbsn} 如果陈述以上为真，下列哪一项也一定为真？ \end{CJK*} \\
        &&   \begin{CJK*}{UTF8}{gbsn} (A)有些律师不是运动员。 \end{CJK*} \\
        &&   \begin{CJK*}{UTF8}{gbsn} (B)有些运动员不是律师。 \end{CJK*} \\
        &&   \begin{CJK*}{UTF8}{gbsn} (C)没有律师是运动员。 \end{CJK*} \\
        &&   \begin{CJK*}{UTF8}{gbsn} (D)所有运动员都是银行家。 \end{CJK*} \\
        &&   \begin{CJK*}{UTF8}{gbsn} 答案: (B)有些运动员不是律师。 \end{CJK*} \\

        && \textbf{Indonesian Version:} \\
        && \emph{Isi: Pada pertemuan yang dihadiri oleh para bankir, atlet, dan pengacara, semua bankir adalah atlet dan tidak ada} \\
        &&  \emph{satupun pengacara yang merupakan bankir.} \\
        && \emph{Pertanyaan: Jika pernyataan di atas benar, manakah pernyataan berikut yang juga benar?} \\
        && \emph{(A) Beberapa pengacara bukanlah atlet.} \\
        && \emph{(B) Beberapa atlet bukan pengacara.} \\
        && \emph{(C) Tidak ada pengacara yang merupakan atlet.} \\
        && \emph{(D) Semua atletnya adalah bankir.} \\
        && \emph{Jawaban: (B) Beberapa atlet bukan pengacara.} \\
        && \textbf{Same sample in Spain, Vietnamese and Malay languages ...} \\
        
        \bottomrule
        \end{tabular}
        \end{adjustbox}
        \caption{
            An example of the instructions and samples from our newly proposed datasets. One instruction is sampled for each dataset.
        }   
        \label{tab:seaeval_examples}
    \end{table*}

    \begin{table*}[t]
        \centering
        \begin{adjustbox}{width=0.98\textwidth,center}
        \begin{tabular}{ l | p{5.5cm} | p{5.5cm} | p{5.5cm} | p{5.5cm} }
        \toprule
         \textbf{Language} & \multicolumn{1}{c|}{\textbf{Korean}} & \multicolumn{1}{c|}{\textbf{Malay}} & \multicolumn{1}{c|}{\textbf{Turkish}} & \multicolumn{1}{c}{\textbf{German}} 
         \\ \midrule
         
         \textbf{Question}
         & \begin{CJK}{UTF8}{mj}다음 문제에서 올바른 답을 선택해 주세요.\end{CJK}
         & \emph{Sila pilih jawapan yang betul untuk soalan berikut.}
         & \emph{Lütfen aşağıdaki soru için doğru cevabı seçin.}
         & \emph{Bitte wählen Sie die richtige Antwort für die}
         \\ 
         
         & \begin{CJK}{UTF8}{mj}흰 빛이 프리즘을 통과할 때, 초록보다\end{CJK}
         & \emph{Apabila cahaya putih melalui prisma, cahaya yang}
         & \emph{Beyaz ışık bir prizmadan geçtiğinde yeşilden}
         & \emph{folgende Frage. Welches Licht wird stärker}
         \\
         
         & \begin{CJK}{UTF8}{mj}더 많이 굴절되는 빛은 무엇입니까?\end{CJK}
         & \emph{membengkok lebih daripada hijau ialah}
         & \emph{daha fazla kırılan ışık}
         & \emph{gebeugt als grünes Licht wenn weißes Licht}
         \\
         
         & \begin{CJK}{UTF8}{mj}(A) 빨강\end{CJK}
         & \emph{(A) Merah}
         & \emph{(A) kırmızıdır}
         & \emph{durch ein Prisma fällt?}
         \\
         
         & \begin{CJK}{UTF8}{mj}(B) 노랑\end{CJK}
         & \emph{(B) Kuning}
         & \emph{(B) sarıdır}
         & \emph{(A) Rot}
         \\
         
         & \begin{CJK}{UTF8}{mj}(C) 파랑\end{CJK}
         & \emph{(C) Biru}
         & \emph{(C) mavidir}
         & \emph{(B) Gelb}
         \\
         
         & \begin{CJK}{UTF8}{mj}(D) 이 중 어느 것도 아님\end{CJK}
         & \emph{(D) Tiada satu pun daripada ini}
         & \emph{(D) Bunlardan hiçbiri}
         & \emph{(C) Blau}
         \\
         
         &  
         & 
         &
         & \emph{(D) Keine von diesen}
         \\

         \midrule
         
         \textbf{Answer} 
         & \begin{CJK}{UTF8}{mj}(C) 파랑\end{CJK}
         & \emph{(A) Merah}
         & \emph{(C) mavidir}
         & \emph{(A) Rot}
         \\

         & \emph{\textbf{In English}: (C) Blue }
         & \emph{\textbf{In English}: (A) Red }
         & \emph{\textbf{In English}: (C) Blue }
         & \emph{\textbf{In English}: (A) Red }
         \\


         \midrule

         \textbf{Correctness}
         & \multicolumn{1}{c|}{\textcolor{ao(english)}{\cmark}} 
         & \multicolumn{1}{c|}{\textcolor{red}{\xmark}} 
         & \multicolumn{1}{c|}{\textcolor{ao(english)}{\cmark}} 
         & \multicolumn{1}{c}{\textcolor{red}{\xmark}} 
         \\
         \bottomrule
        \end{tabular}
        \end{adjustbox}
        \caption{
            An example from our Cross-MMLU dataset on Korean, Malay, Turkish and German
        }

        \label{tab:cross-lingual-alignment-example-malay-korean}
    \end{table*}

\end{document}